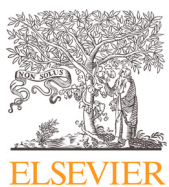

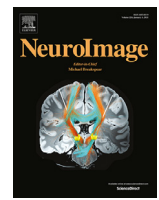

# Deep fiber clustering: Anatomically informed fiber clustering with self-supervised deep learning for fast and effective tractography parcellation

Yuqian Chen [a,b], Chaoyi Zhang [b], Tengfei Xue [a,b], Yang Song [c], Nikos Makris [a], Yogesh Rathi [a], Weidong Cai [b], Fan Zhang [a,*], Lauren J. O'Donnell [a,*]

[a] Harvard Medical School, MA, USA
[b] The University of Sydney, NSW, Australia
[c] The University of New South Wales, NSW, Australia

## ARTICLE INFO

## ABSTRACT

White matter fiber clustering is an important strategy for white matter parcellation, which enables quantitative analysis of brain connections in health and disease. In combination with expert neuroanatomical labeling, data-driven white matter fiber clustering is a powerful tool for creating atlases that can model white matter anatomy across individuals. While widely used fiber clustering approaches have shown good performance using classical unsupervised machine learning techniques, recent advances in deep learning reveal a promising direction toward fast and effective fiber clustering. In this work, we propose a novel deep learning framework for white matter fiber clustering, Deep Fiber Clustering (DFC), which solves the unsupervised clustering problem as a self-supervised learning task with a domain-specific pretext task to predict pairwise fiber distances. This process learns a high-dimensional embedding feature representation for each fiber, regardless of the order of fiber points reconstructed during tractography. We design a novel network architecture that represents input fibers as point clouds and allows the incorporation of additional sources of input information from gray matter parcellation. Thus, DFC makes use of combined information about white matter fiber geometry and gray matter anatomy to improve the anatomical coherence of fiber clusters. In addition, DFC conducts outlier removal naturally by rejecting fibers with low cluster assignment probability. We evaluate DFC on three independently acquired cohorts, including data from 220 individuals across genders, ages (young and elderly adults), and different health conditions (healthy control and multiple neuropsychiatric disorders). We compare DFC to several state-of-the-art white matter fiber clustering algorithms. Experimental results demonstrate superior performance of DFC in terms of cluster compactness, generalization ability, anatomical coherence, and computational efficiency.

## 1. Introduction

Diffusion magnetic resonance imaging (dMRI) tractography is an advanced imaging technique that uniquely enables in vivo mapping of the brain's white matter connections at macro scale (Basser et al., 2000; Mori et al., 1999). Tractography enables quantitative analysis of the brain's structural connectivity in many applications such as neurological development, aging, and brain disease (Ciccarelli et al., 2008; Essayed et al., 2017; Piper et al., 2014; Yamada et al., 2009; Zhang et al., 2022b). However, when performing whole brain tractography, hundreds of thousands to millions of fibers (or streamlines)[1] are generated, which are not directly useful to clinicians or researchers. Therefore, to enable fiber tract quantification and visualization, it is essential to perform tractography parcellation where the massive number of tractography fibers is divided into multiple subdivisions (Zhang et al., 2022b).

### 1.1. Tractography parcellation methods

Two popular categories of tractography parcellation methods (O'Donnell et al., 2013; Zhang et al., 2022b) include cortical-parcellation-based methods that group fibers according to their endpoints in gray matter regions (Gong et al., 2009), and white matter fiber clustering methods that group fibers with similar geometric trajectories (Brun et al., 2004; Chekir et al., 2014; Garyfallidis et al., 2018;

---

Guevara et al., 2012; Li et al., 2010; O'Donnell et al., 2013; Román et al., 2017; Siless et al., 2018; St-Onge et al., 2021; Tunç et al., 2014; Vázquez et al., 2020a; Wu et al., 2020; Yoo et al., 2015). Compared to cortical-parcellation-based methods, white matter fiber clustering methods can obtain more consistent parcellations across subjects (Sydnor et al., 2018; Zhang et al., 2017a, 2018c) and demonstrate higher test-retest reproducibility (Zhang et al., 2019b). White matter fiber clustering enables studies of the brain's white matter across the lifespan in health and disease (Cousineau et al., 2017; Ji et al., 2019; Maier-Hein et al., 2017; O'Donnell et al., 2017; Prasad et al., 2014; Tunç et al., 2016; Zekelman et al., 2022; Zhang et al., 2018a). White matter fiber clustering also enables the creation of tractography atlases and the study of white matter anatomy (Battocchio et al., 2022; Guevara et al., 2022, 2020; Levitt et al., 2021; O'Donnell et al., 2017; Lauren J. O'Donnell and Westin, 2007; Román et al., 2022, 2021, 2017; Tunç et al., 2016, 2014; Vázquez et al., 2020a; Yeh et al., 2018; Zhang et al., 2018c). A popular strategy for the creation of white matter tractography atlases incorporates machine learning fiber clustering methods to automatically group streamlines, followed by expert neuroanatomical annotation of fiber clusters to define anatomical structures (Yeh et al., 2018) as well as false positive connections (Zhang et al., 2018c). The improvement of fiber clustering algorithms can enhance the depiction of understudied regions, such as the superficial white matter (Román et al., 2022; Xue et al., 2023) or the cerebellum (Zhang et al., 2018c), and can enable the automated study of very large datasets (Zhang et al., 2022a).

Many methods have been proposed for white matter fiber clustering (see (Zhang et al., 2022b) for a review of methods). Generally, fiber clustering methods compute distances between fibers and then group fibers into clusters using computational clustering methods. Several methods have been designed for rapid clustering of tractography from an individual subject, e.g., to create a compact representation of whole-brain tractography for further processing (Garyfallidis et al., 2016, 2012; Guevara et al., 2011; Vázquez et al., 2020b). For example, QuickBundles employs the minimum average direct-flip fiber distance with a fast linear-time clustering algorithm (Garyfallidis et al., 2012), while FFClust first clusters fiber points and then groups fibers into compact clusters with high efficiency (Vázquez et al., 2020b). Other fiber clustering methods cluster tractography from multiple subjects in a groupwise fashion (O'Donnell and Westin, 2005) to create population-based tractography atlases (O'Donnell et al., 2017; Lauren J. O'Donnell and Westin, 2007; Tunç et al., 2016, 2014; Zhang et al., 2018c). For example, WhiteMatterAnalysis uses the mean distance between pairs of closest fiber points to enable groupwise spectral clustering (Zhang et al., 2018c). Finally, other white matter fiber clustering methods use information from an anatomical parcellation of the brain. In an early approach, anatomical information from a white matter parcellation was used to guide the clustering of fiber tracts (Maddah et al., 2008). More recently, "connectivity-driven" fiber clustering is based on the connectivity of the voxels through which fibers pass (Tunç et al., 2016, 2014, 2013), and AnatomiCuts clusters fibers based on their position relative to anatomical regions (Siless et al., 2020, 2018).

Though existing white matter fiber clustering methods have shown good performance, several key challenges remain. First, it is computationally expensive to calculate all pairwise fiber similarities considering the large number of fibers in whole brain tractography. Second, the computed fiber similarities can be sensitive to the order of points along the fibers. This is a problem because a fiber can equivalently start from either end (Garyfallidis et al., 2012; Zhang et al., 2020). Third, false positive fibers are prevalent in tractography and outliers may exist in obtained fiber clusters (Legarreta et al., 2021; Maier-Hein et al., 2017). Therefore, outlier removal methods are needed to remove undesired fibers from cluster results. Fourth, current methods mostly use descriptions of either white matter fiber geometry (i.e., fiber point spatial coordinates (Brun et al., 2004; Chen et al., 2021; Corouge et al., 2004a; Garyfallidis et al., 2012; Ngattai Lam et al., 2018; Vázquez et al., 2020b; Zhang et al., 2018c) or gray matter anatomical parcellation (i.e., corti-cal and subcortical segmentations (Siless et al., 2018)) for fiber clustering. It is a challenge to combine both white matter fiber geometry and gray matter anatomical parcellation information to improve the clustering results. Finally, it is important to identify cluster correspondences across subjects for group-wise analysis. To achieve this goal, some studies perform fiber clustering across subjects to form an atlas and predict clusters of new subjects with correspondence to the atlas (Lauren J. O'Donnell and Westin, 2007; Tunç et al., 2014; Zhang et al., 2018c), while other approaches first perform within-subject fiber clustering and then match the fiber clusters across subjects (Garyfallidis et al., 2012; Guevara et al., 2012; Huerta et al., 2020; Siless et al., 2020, 2018).

## 1.2. Unsupervised feature learning and clustering

In recent years, deep learning has demonstrated superior performance in computer vision tasks such as object classification, detection and segmentation (He et al., 2017; Ronneberger et al., 2015; Simonyan and Zisserman, 2014). Deep-learning-based clustering has also been extensively studied as an unsupervised learning task (Károly et al., 2018). An intuitive way to perform unsupervised deep clustering is to extract feature embeddings with neural networks and then perform clustering on these embeddings to form clusters. The learned embeddings are high-level representations of input data and have been shown to be informative for downstream tasks (Song et al., 2016), such as clustering (Tian et al., 2014; Xie et al., 2016). Autoencoder networks are widely used to learn unsupervised feature embeddings because they do not require ground truth labels (Guo et al., 2017; Xie et al., 2016). The representative work is the Deep Embedded Clustering framework, which performs simultaneous embedding of input data and cluster assignments in an end-to-end way (Xie et al., 2016). Deep Convolutional Embedded Clustering (DCEC) extends Deep Embedded Clustering from 1D feature vector clustering to 2D image clustering (Guo et al., 2017).

Another promising approach for learning feature embeddings is self-supervised learning, which is a subclass of unsupervised learning that shows advanced performance in many applications (Kolesnikov et al., 2019; van den Oord et al., 2018). Deep embeddings are obtained by designing a pretext task such as predicting context (Doersch et al., 2015) or image rotation (Komodakis and Gidaris, 2018) and generating pseudo labels from the input data to guide network training, without involving any manual annotations. The learned feature embeddings (usually referred to as the high-level feature representations) can then be transferred to downstream tasks such as clustering.

Recently, attempts have been made to apply supervised deep learning approaches for tractography segmentation (Gupta et al., 2017a; Gupta et al., 2017b; Liu et al., 2019, 2022; Ngattai Lam et al., 2018; Wasserthal et al., 2018; Xue et al., 2022; Xu et al., 2019; Zhang et al., 2020). In these studies, fibers from the whole brain are classified into anatomically meaningful fiber tracts based on labeled training datasets. To alleviate the requirement of ground truth labels, one recently proposed method (Xu et al., 2021) has shown the potential of using unsupervised deep learning for fiber clustering; however, it requires complex feature extraction procedures to generate inputs of the neural network. We proposed a novel unsupervised deep learning framework in our MICCAI work (Chen et al., 2021), where we adopted self-supervised learning to achieve fast and effective white matter fiber clustering. However, it also requires an extra step to generate inputs of the neural network (FiberMaps (Zhang et al., 2020)) from the fiber points.

In tractography data, each fiber is encoded as a set of points along its trajectory. Therefore, it could be intuitive and efficient to represent and process fiber data as point clouds, which are an important geometric data format. In addition, each fiber could naturally be represented as a graph, where points are considered to be nodes. In these ways, original fiber point coordinates could be processed directly with point-based neural networks or Graph Neural Networks, which have demonstrated successful applications in geometric data processing (Chen et al., 2017;

Qi et al., 2017; Welling and Kipf, 2016). Another benefit of these representations for tractography data is that the point cloud or graph representation of a fiber is not sensitive to the point ordering along the fiber. In recent studies, fibers have been represented as point clouds for tractography-related supervised learning tasks (Astolfi et al., 2020; Chen et al., 2022; Logiraj et al., 2021; Xue et al., 2023), contributing to superior performance and high efficiency. In the computer vision community, unsupervised point cloud and graph clustering have been achieved in several studies by learning representations of inputs first and then performing traditional clustering on learned embeddings (Hassani and Haley, 2019; Tian et al., 2014). However, we have found no related work using point clouds or graphs for unsupervised white matter fiber clustering tasks yet.

### 1.3. Contributions

In this study, we propose a novel deep learning framework for fast and effective white matter fiber clustering. The whole framework is trained in an end-to-end way with fiber point coordinates as inputs and cluster assignments of fibers as outputs. Using a point cloud representation of input fibers, our framework learns deep embeddings by pretraining the neural network in a self-supervised manner and then fine-tunes the network in a self-training manner (Xie et al., 2016) with the task of updating cluster assignments. At the inference stage, the trained fiber clustering pipeline can be applied to parcellate independently acquired datasets.

This paper has five contributions. First, input fibers are represented as point clouds, which are compact representations and improve efficiency via adopting point-based neural networks. Second, self-supervised learning is adopted in our pipeline with a designed pretext task to obtain feature embeddings insensitive to fiber point ordering for input fibers, enabling subsequent clustering. Third, white matter fiber geometric information and gray matter anatomical parcellation information are combined in the proposed framework to obtain spatially compact and anatomically coherent clusters. Fourth, outliers are removed after cluster assignment by rejecting fibers with low soft label assignment probabilities. Fifth, our approach automatically creates a multi-subject fiber cluster atlas that is applied for white matter parcellation of new subjects.

The preliminary version of this work, referred to as DFC$_{conf}$, was published in MICCAI 2021 (Chen et al., 2021). In this paper, we extend our previous work by: 1) adopting a new fiber representation (i.e., point cloud), with a comprehensive evaluation of different representations of tractography data including point clouds, graphs, and images; 2) adding cortical surface parcellation information in addition to anatomical region information to further improve cluster anatomical coherence; 3) a new cluster-adaptive outlier removal process to filter anatomically implausible fibers while maintaining good generalization across subjects; and 4) demonstrating the robustness of our method on additional datasets with different acquisitions, ages, and health conditions.

## 2. Methods

The overall pipeline of DFC is shown in Fig. 1. The training process includes two stages: pretraining and clustering. In the pretraining stage, neural networks are trained to perform a self-supervised pretext task and obtain feature embeddings of a pair of input fibers (point clouds), followed by k-means clustering (Likas et al., 2003) to obtain initial clusters. In the clustering stage, based on the neural network initialized in the pretraining stage, clustering results are fine-tuned via a

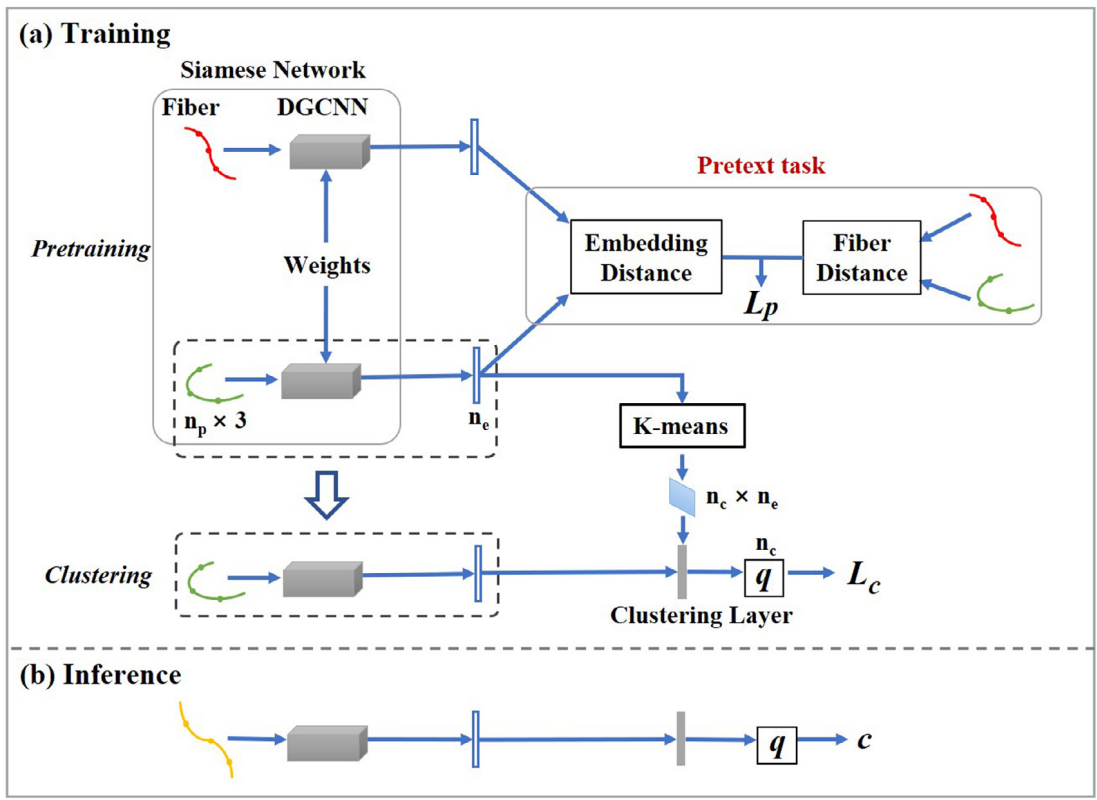

**Fig. 1.** Overview of our DFC framework. A self-supervised learning strategy is adopted with the pretext task of pairwise fiber distance prediction. In the pretraining stage, input fibers are encoded as embeddings with the Siamese Networks. K-means clustering is then performed on the obtained embeddings to generate initial cluster centroids. In the clustering stage, based on the neural network of the pretraining stage, a clustering layer is connected to the embedding layer and generates soft label assignment probabilities $q$ (as shown in the orange dashed box). During training, a prediction loss ($L_p$) and a KL divergence loss ($L_c$) are combined for network optimization. During inference, an input fiber is assigned to cluster $c$ with the maximum soft label assignment probability, which is calculated from the trained neural network. ($n_p$: number of points; $n_e$: dimension of embeddings; $n_c$: number of clusters).

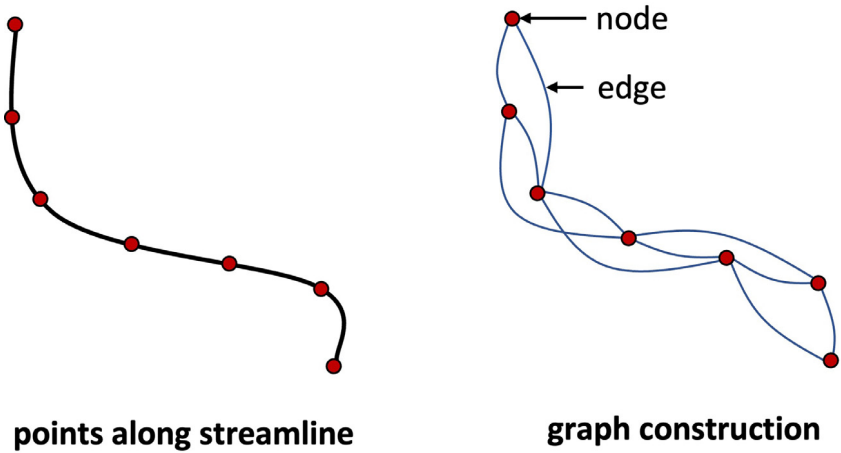

**Fig. 2.** Illustration of the process of graph construction for the input of the DGCNN model.

**points along streamline**

**graph construction**

node

edge

self-training manner (Xie et al., 2016). This process is done by adding a clustering layer (see details in Section 2.3) where cluster assignment probabilities are calculated from the distances between feature embeddings and cluster centroids. In this way, for each input fiber, the output is a probability vector with a dimension of the number of clusters. During inference, for each fiber represented as a point cloud, an embedding is predicted by the trained neural network, and the fiber is assigned to the closest cluster by calculating the distances between its embedding and all cluster centroids. By performing cluster assignments with the trained neural network, our method automatically achieves cluster correspondence across subjects.

### 2.1. Input fiber geometry and anatomical information

In this work, we adopt point clouds as representations of fibers. Considering that the neighborhood relationship among points along a fiber could provide contextual information for clustering, we adopt the Dynamic Graph Convolutional Neural Network (DGCNN) model (Wang et al., 2019). The DGCNN model contains an edge feature engineering module, EdgeConv, which was proposed to capture the local geometric structure formed by points and their neighbors. In a similar way to DGCNN, a graph is constructed for each fiber with nodes representing fiber points and edges built between nearby points along the fiber (as illustrated in Fig. 2). Considering that fiber points are distributed along a fiber, we construct a graph with edges connecting each set of $k$ ($k = 4$ in this study) nearest points along a fiber (instead of edges connecting $k$ nearby points based on Euclidean distance as in the original DGCNN method) (Astolfi et al., 2020). We note that while the graph structure for all fibers is the same, the node features (spatial coordinates of fiber points) of each graph are different so that fibers belonging to different clusters can be distinguished. The inputs to the DGCNN model are point clouds with dimension $n_p \times 3$, where $n_p$ is the number of fiber points and 3 is the number of spatial coordinates of fiber points.

To provide anatomical context to improve performance at the fiber clustering stage (Section 2.3), we augment the white matter fiber geometry information with gray matter anatomical parcellation information. This information includes anatomical regions and cortical parcellations obtained from Freesurfer (Fischl, 2012) using the Desikan-Killiany Atlas (Desikan et al., 2006). To describe the anatomical regions through which each fiber passes, each point in a fiber is assigned the label of the anatomical region it intersects. Similarly, fiber endpoints are associated with the cortical parcellation label of the closest point on the cortical surface.

### 2.2. Pretraining with self-supervised deep embedding

In the pretraining stage, we propose a novel self-supervised learning approach to obtain deep embeddings of fibers. A pretext task is designed to obtain pairs of embeddings with distances similar to their corresponding fiber distances, enabling subsequent clustering in embedding space. Specifically, the pretext task is to predict the distance between a pair of input fibers, where their self-supervised pseudo label is given as the pairwise fiber distance between their pointwise spatial coordinates. To calculate the fiber distance, we adopt the minimum average direct-flip distance, which is widely applied in white matter fiber clustering (Garyfallidis et al., 2012; Zhang et al., 2018c). This fiber distance considers the order of points along the fibers, and it remains the same when a fiber is equivalently represented starting from either endpoint. With fiber distances as pseudo labels, the network is guided to produce similar embeddings for similar fibers, even in the presence of flipped fiber point orderings.

To perform the pretext task of fiber distance prediction, we adopt a Siamese Network (Chopra et al., 2005), which has two subnetworks with shared weights. Generally, a pair of inputs is put into the subnetworks, respectively, and a pair of deep embeddings is generated from the subnetworks. In this work, a pair of fibers (point cloud sets) is used as the input to the point-cloud-based neural network. We employ DGCNNs as subnetworks of the Siamese Network to obtain feature embeddings. Each DGCNN subnetwork is composed of 5 EdgeConv layers followed by 3 fully connected layers. The subnetworks output a pair of deep embeddings corresponding to the input pair.

In the general use of Siamese Network, a fully connected layer follows the subnetworks and outputs a similarity score. In our work, we replace the last fully connected layer with a direct calculation of the pairwise Euclidean distance between the learned deep embeddings. The mean squared error between the predicted distance and fiber distance (pseudo label) is calculated as the distance prediction loss $L_p$.

### 2.3. Clustering integrating anatomical information

After the pretraining stage, the weights of the Siamese Network are initialized with the pretrained weights, and initial clusters are obtained by performing k-means clustering (Likas et al., 2003) on the generated embeddings. The clustering stage of our method is developed from the Deep Convolutional Embedded Clustering model (Guo et al., 2017). Following the DGCNN architecture, a clustering layer is designed to encapsulate cluster centroids as its trainable weights and compute soft label assignment probabilities $q_{ij}$ using Student's t-distribution (Maaten, 2008):

$$q_{ij} = \frac{\left(1 + \left\| z_i - \mu_j \right\|^2\right)^{-1}}{\sum_{j'} \left(1 + \left\| z_i - \mu_{j'} \right\|^2\right)^{-1}} \tag{1}$$

where $z_i$ is the embedding of fiber $i$ and $\mu_j$ is the centroid of cluster $j$ (note that $z_i$ and $\mu_j$ have the same dimensionality). $q_{ij}$ is the probability of assigning fiber $i$ to cluster $j$. The network is trained in a self-training manner and its clustering loss $L_c$ is defined as a KL divergence loss (Xie et al., 2016). The distance prediction loss is retained in the clustering stage, and the total loss is $L = L_p + \lambda L_c$, where $\lambda$ is the weight of $L_c$. During inference, a fiber $i$ is assigned to the cluster with the maximum $q_{ij}$ referred to as $q_m$.

We improve the clustering stage described above by incorporating gray matter anatomical parcellation information into the neural network. We design a new soft label assignment probability definition, which extends (1) to encourage grouping of fibers that pass through the same anatomical regions and cortical parcels:

$$q_{ij} = \frac{\left(1 + \left\|z_i - \mu_j\right\|^2 * \left(1 - D_{ij}^a\right) * \left(1 - D_{ij}^c\right)\right)}{\sum_{j'} \left(1 + \left\|z_i - \mu_{j'}\right\|^2 * \left(1 - D_{ij'}^a\right)\left(1 - D_{ij'}^c\right)\right)^{-1}} \qquad (2)$$

where is the Dice score between the set of anatomical regions passed through by fiber $i$ and those passed through by cluster $j$. To define this set of anatomical regions, we use the *tract anatomical profile* method that includes regions intersected by over 40% of fibers as in (Garyfallidis et al., 2012; Zhang et al., 2018c). Similarly, $D_{ij}^c$ quantifies the agreement between the set of cortical regions intersected by the endpoints of fiber $i$ and those intersected by the endpoints of cluster $j$. $D_{ij}^c$ is defined as the percentage of endpoints in cluster $j$ that are within the cortical regions intersected by the endpoints of fiber $i$. Analogous to the tract anatomical profile, we propose to call the percentage of endpoints within each intersected cortical region the *tract surface profile*. During training, the tract anatomical profile and tract surface profile are initially calculated from the clusters generated by $k$-means and updated iteratively with new predictions during the clustering stage. During inference, soft label assignments are calculated using (2).

### 2.4. Cluster-adaptive outlier removal

After initial clustering, outlier fibers may have distinctly different position and shape from most fibers in the cluster, and we empirically found that these outliers often exist in obtained clusters. Therefore, outlier removal is an essential step to filter anatomically implausible fibers (Astolfi et al., 2020; Guevara et al., 2011; Legarreta et al., 2021; Mendoza et al., 2021; Zhang et al., 2018c). In our previous work (Chen et al., 2021), we removed outliers by directly rejecting fibers with a label assignment probability $q_m$ lower than an absolute threshold. This method could potentially remove plausible fibers, as it ignores the variability of $q_m$ across clusters with different anatomy.

Therefore, we propose a novel cluster-adaptive outlier removal method. It is also based on the maximum label assignment probability $q_m$, considering that fibers with higher $q_m$ tend to have higher confidence of belonging to the corresponding clusters and are thus less likely to be outliers. In our proposed method, fibers with low soft label assignment probabilities are removed based on a cluster-specific threshold, rather than an absolute threshold across all clusters. Specifically, for each cluster ($c$), we calculate the mean ($m_c$) and the standard deviation ($s_c$) of the label assignment probabilities of all fibers assigned to this cluster. Then, a threshold is computed as $T_c = m_c - n * s_c$ such that any fiber with a label assignment probability lower than $T_c$ is removed (where $n$ is a hyperparameter that controls the quantity of removed outlier fibers). The above threshold computation process is commonly used for outlier data detection (Dave and Varma, 2014), and a similar approach was effective in a previous work for fiber clustering (Zhang et al., 2018c).

### 2.5. Implementation details

In the pretraining and clustering stages, our model is trained for 50k iterations with a learning rate of 1e-4 and another 1k iterations with a learning rate of 1e-5. The batch size of training is 1024 and Adam

(Kingma and Ba, 2014) is used for optimization. All methods were tested on a computer equipped with a 2.1 GHz Intel Xeon E5 CPU (8 DIMMs; 32 GB Memory) and an NVIDIA RTX 2080Ti GPU. Deep learning methods were implemented with Pytorch (v1.7.1) (Paszke et al., 2019). The weight of clustering loss $\lambda$ was set to be 0.1, as suggested in (Guo et al., 2017). The source code and the trained model will be made available at https://github.com/SlicerDMRI/DFC.

## 3. Results

### 3.1. Experimental datasets and preprocessing

We used dMRI data from three datasets that were independently acquired from different populations using different imaging protocols and scanners, as shown in Table 1. Data of 100 subjects from the Human Connectome Project (HCP) (Van Essen et al., 2013) were used for model training, and another 120 subjects from HCP, Consortium for Neuropsychiatric Phenomics (CNP) (Poldrack et al., 2016) and Parkinson's Progression Markers Initiative (PPMI) (Marek et al., 2011) (across genders, ages, different health conditions, and different acquisitions) were used for testing.

For each subject, whole-brain tractography was performed using a two-tensor unscented Kalman filter method (Malcolm et al., 2010; Reddy and Rathi, 2016). Fibers shorter than 40 mm were removed to avoid any bias toward implausible short fibers (Guevara et al., 2012; Jin et al., 2014). The average numbers of fibers per subject obtained with the whole-brain tractography were approximately 490,000 for the HCP dataset, 950,000 for the PPMI dataset, and 880,000 for the CNP dataset. All tractography data were co-registered using a tractography-based registration method (O'Donnell et al., 2012). In order to obtain gray matter anatomical parcellation information (i.e., the anatomical regions each fiber passed through and the cortical regions each fiber connected to), we performed Freesurfer parcellation (Fischl, 2012) on the T1w data, which was then registered to the dMRI data. (Note: for HCP data, we used the provided FreeSurfer parcellation that had been co-registered with the dMRI data; for the CNP and PPMI data, we performed a non-linear registration using ANTs (Avants et al., 2009).)

During model training, 10,000 fibers were randomly selected from each of the 100 training subjects, generating a training dataset of 1 million fiber samples. During the pretraining stage, each fiber sample was paired with another randomly selected sample other than itself, generating 1 million fiber pairs, to learn the embedding features. During the training stage, the training dataset was parcellated into 800 clusters, resulting in an average of 1250 fibers in each cluster. Then, the trained model was applied to the whole-brain tractography of each testing subject for subject-specific white matter fiber clustering. For fast and efficient processing of the large number of fiber samples during model training and inference, fibers were downsampled to $n_p$ points before being input into the network. In this study, we set $n_p$ as 14 because this number enables good performance with relatively low computational costs

**Table 1**

Demographics and dMRI acquisition of the three independent datasets tested. (Abbreviations: Y - years old; F - female; M - male; H - healthy; SZ - schizophrenia; ADHD - attention-deficit/hyperactivity disorder; BP - bipolar disorder; PD - Parkinson's disease.)

| Dataset | N | Demographics | dMRI data |
|---|---|---|---|
| HCP | 50 | 22 to 35 Y; F: 32, M: 18; H: 50 | $b$ = 3000 s/mm²; 108 directions; TE/TR=89/5520 ms; resolution=1.25 mm isotropic |
| CNP | 40 | 21 to 50 Y; F: 17, M: 23; H: 11, SZ: 12, BP: 12, ADHD: 5 | $b$ = 1000s/mm²; 64 directions; TE/TR = 93/9000 ms; resolution = 2 mm isotropic |
| PPMI | 30 | 51 to 75 Y; F: 9, M: 21; H: 14, PD: 16 | $b$ = 1000s/mm²; 64 directions; TE/TR = 88/7600 ms; resolution = 2 mm isotropic |

in terms of inference time and memory usage (for details see Supplementary Material 1). All anatomical region labels (from all fiber points) were preserved for input into the network without any downsampling.

### 3.2. Experimental metrics

We adopted four metrics to quantitatively evaluate white matter fiber clustering results. These metrics enable evaluation of the quality of a white matter tractography parcellation from several perspectives.

#### 3.2.1. Davies-Bouldin (DB) index

DB index is a commonly used metric in unsupervised clustering tasks (Xu and Tian, 2015), and it has been recently adopted for fiber clustering evaluation (Vázquez et al., 2020b). It simultaneously measures within-cluster scatter and between-cluster separation, as the ratio of intra- and inter-cluster fiber distances $DB = (1/n)\sum_n^{k=1} \max_{i \neq j}((\alpha_i + \alpha_j)/d(c_i, c_j))$, where $n$ is the number of clusters, $\alpha_i$ and $\alpha_j$ are mean intra-cluster distances, and $d(c_i, c_j)$ is inter-cluster distance (minimum average directflip distance between centroids $c_i$ and $c_j$ of cluster $i$ and $j$, where the centroid is defined as the fiber with minimum average distance to all other fibers in the cluster). A smaller DB index indicates better separation between clusters.

#### 3.2.2. White Matter Parcellation Generalization (WMPG)

WMPG measures the percentage of successfully detected clusters in an individual subject (Zhang et al., 2018c). Clusters with over 20 fibers are considered to be successfully detected (Zhang et al., 2018c).

#### 3.2.3. Tract Anatomical Profile Coherence (TAPC)

This metric measures if fibers within the same cluster pass through the same anatomical regions (Zhang et al., 2018c). It is calculated as the Dice score between each fiber's intersected anatomical regions and its assigned cluster's anatomical regions (i.e. the tract anatomical profile of the cluster (Section 2.3), where a high value suggests a high anatomical region coherence of the cluster. The TAPC of a cluster is calculated as the mean of Dice scores across all fibers within the cluster, and the TAPC score of a subject is computed as the mean TAPC of all clusters.

#### 3.2.4. Tract Surface Profile Coherence (TSPC)

We propose a new metric, TSPC, to evaluate the coherence of cortical terminations of fibers within a cluster. The TSPC is defined as the average tract surface profile (Section 2.3) across the cortical regions intersected by fiber endpoints within the cluster. A higher TSPC indicates that fibers within a cluster terminate in a smaller set of cortical parcels. The TSPC of a subject is computed as the mean TSPC of all clusters.

### 3.3. Experiments and results

We performed five experimental evaluations, including 1) comparison to state-of-the-art 2) comparison to baseline Deep Convolutional Embedded Clustering, 3) ablation study, 4) evaluation of input representations and network architectures, and 5) evaluation of outlier fiber removal. Experiment results of 1) and 2) are reported using the three testing datasets and those of 3), 4) and 5) are reported using the HCP testing dataset.

#### 3.3.1. Comparison to state-of-the-art methods

We compared the proposed DFC with three state-of-the-art methods: WhiteMatterAnalysis (Zhang et al., 2018c), QuickBundles (Garyfallidis et al., 2012) and DFC$_{conf}$ (Chen et al., 2021). WhiteMatterAnalysis is an atlas-based white matter fiber clustering method that shows good performance and strong correspondence across subjects. QuickBundles is a widely used white matter fiber clustering method that performs clustering on each subject and achieves group correspondence with post-processing steps. We used open-source software packages WhiteMatterAnalysis v0.3.0 (github.com/SlicerDMRI/

whitematteranalysis) and Dipy v1.3.0 (dipy.org) with default settings to implement WhiteMatterAnalysis and QuickBundles, respectively. DFC$_{conf}$ is the preliminary version of this work that adopts FiberMap (Zhang et al., 2020), which is a 2D multi-channel feature descriptor that encodes spatial coordinates of points along each fiber, as representation of input fibers. Cluster correspondence across subjects is automatically generated by DFC and DFC$_{conf}$. For each method, we performed white matter fiber clustering to output 800 clusters, which has been suggested to be a good whole brain tractography parcellation scale (Wu et al., 2021; Zhang et al., 2018c). (For QuickBundles, a number as close as possible to 800 clusters was initially obtained by tuning parameters. For each dataset, the obtained numbers of clusters were $822.64 \pm 13.01$ for HCP, $855.13 \pm 32.62$ for PPMI and $851.31 \pm 31.58$ for CNP. Next, post-processing was performed to identify corresponding QuickBundles clusters across subjects by finding the closest cluster of each subject to each cluster of a template subject as in (Garyfallidis et al., 2012). This resulted in 800 QuickBundles clusters for all subjects, the same number of clusters as the template subject.) For fair comparison, we adjusted the outlier removal threshold so that DFC and DFC$_{conf}$ removed approximately the same percentage of fibers (25.5%−26.5% for HCP, 34%−36% for CNP, and 34.5%−35.5% for PPMI) as WhiteMatterAnalysis. The outlier removal threshold $n$ was set to 0.7, 0.83, and 0.85 for HCP, CNP and PPMI respectively in DFC and 0.0185, 0.018 and 0.018 respectively in DFC$_{conf}$. Quantitative comparison across these methods was performed using the evaluation metrics introduced in Section 3.2. Statistical analysis was performed for the results of each metric using a one-way ANOVA followed by post hoc pairwise comparisons using paired t-tests.

Table 2 gives the quantitative results of the state-of-the-art comparison experiment in the three testing datasets. Note that this experiment employs the DB, WMPG, TAPC and TSPC metrics (defined in Section 3.2) that measure cluster compactness, cluster generalization across subjects, cluster anatomical coherence, and cluster cortical anatomical coherence, respectively. Overall, for each metric, the ANOVA analyses show there are significant differences among the compared methods in each testing dataset ($p < 0.0001$ in all analyses). For the DB and TSPC metrics, post hoc paired t-tests show that the proposed DFC obtains significantly better results than all other methods in all testing datasets. For the WMPG metric, DFC obtains significantly better results than all other methods, except in the CNP and PPMI datasets where DFC and WhiteMatterAnalysis have no significant differences. For the TAPC metric, the proposed DFC method is significantly better than the WhiteMatterAnalysis and QuickBundles methods, though our conference version DFC$_{conf}$ achieves the best performance.

Next we discuss the quantitative metric results from Table 2 and how they compare across methods. For the DB index, DFC obtains the lowest

**Table 2**

Experimental results on the HCP dataset (50 subjects), CNP dataset (40 subjects) and PPMI dataset (30 subjects). Results are presented as mean value with standard deviation across subjects in parenthesis. The bolded results indicate the best performance for the corresponding evaluation metric. *indicates the metric is significantly different from that of DFC with $p < 0.001$. (Abbreviations: DFC - Deep Fiber Clustering; DFC$_{conf}$ - conference version of Deep Fiber Clustering; WMA - WhiteMatterAnalysis; QB - QuickBundles).

| Methods | | DFC | DFC$_{conf}$ | WMA | QB |
|---------|------|-----|--------------|-----|-----|
| HCP | DB | **2.014 (0.023)** | 2.059 (0.025)* | 2.350 (0.052)* | 2.084 (0.032)* |
| | WMPG | **0.996 (0.004)** | 0.974 (0.010)* | 0.992 (0.008)* | 0.742 (0.025)* |
| | TAPC | 0.844 (0.003) | **0.847 (0.003)** | 0.825 (0.003)* | 0.787 (0.008)* |
| | TSPC | **0.601 (0.008)** | 0.564 (0.008)* | 0.526 (0.007)* | 0.472 (0.018)* |
| CNP | DB | **2.127 (0.027)** | 2.199 (0.029)* | 2.351 (0.034)* | 2.163 (0.042)* |
| | WMPG | 0.970 (0.015) | 0.939 (0.022)* | **0.971 (0.014)** | 0.810 (0.022)* |
| | TAPC | 0.830 (0.002) | **0.836 (0.004)** | 0.815 (0.003)* | 0.758 (0.009)* |
| | TSPC | **0.498 (0.007)** | 0.452 (0.010)* | 0.458 (0.008)* | 0.361 (0.019)* |
| PPMI | DB | **2.119 (0.028)** | 2.200 (0.031)* | 2.322 (0.032)* | 2.162 (0.053)* |
| | WMPG | **0.978 (0.014)** | 0.944 (0.027)* | 0.977 (0.014) | 0.829 (0.032)* |
| | TAPC | 0.832 (0.003) | **0.837 (0.004)** | 0.819 (0.003)* | 0.756 (0.013)* |
| | TSPC | **0.476 (0.009)** | 0.432 (0.011)* | 0.436 (0.008)* | 0.339 (0.026)* |

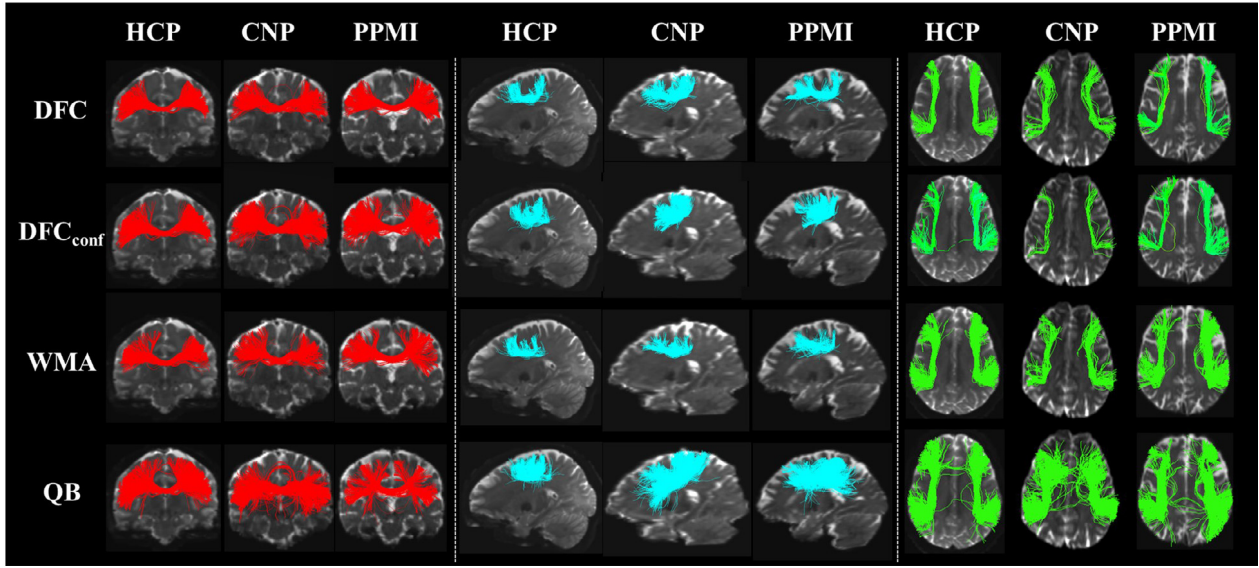

**Fig. 3.** Visualization of example clusters from four methods (DFC, DFC$_{conf}$, WMA, QB) across three datasets (HCP, CNP, PPMI). Three example clusters are selected within known CC4, Sup-FP and AF tracts respectively. For clusters within CC4 tracts, an anterior view is displayed; for AF clusters, an inferior view is displayed; for Sup-FP clusters, a left view is displayed. (Abbreviations: CC4 - corpus callosum 4; Sup-FP - superficial-frontal-parietal; AF - arcuate fasciculus; DFC - Deep Fiber Clustering; DFC$_{conf}$ - conference version of Deep Fiber Clustering; WMA - WhiteMatterAnalysis; QB - QuickBundles).

score thus the best performance, while the other methods also obtain relatively similar and low scores, indicating that all compared methods generate compact and well-separated clusters. For WMPG, DFC and WhiteMatterAnalysis obtain the best performance (over 97% of clusters detected in each dataset), followed by DFC$_{conf}$ (over 93% in each dataset), whereas QuickBundles was the least favorable (about 70 to 80% in each dataset). For TAPC, DFC and DFC$_{conf}$ had higher scores, thus obtaining better anatomical region coherence than the other two methods, though DFC$_{conf}$ was slightly better than DFC. For TSPC, DFC obtained the highest score, indicating the best cortical anatomical coherence.

Fig. 3 gives a visualization of clusters obtained from each method. For the purposes of visualization, cluster correspondences were obtained by finding the closest cluster from all comparison methods to each cluster from DFC. (The closest cluster was defined as the cluster with the minimum average pairwise fiber distance to a cluster from DFC. Pairwise fiber distances are often used for measuring distances between clusters or tractograms (Chen et al., 2023; Guevara et al., 2022; Zhang et al., 2018b).) In general, DFC, DFC$_{conf}$ and WhiteMatterAnalysis obtain visually similar clusters, while the clusters from DFC appear to be more compact and anatomically reasonable than those from the other methods. QuickBundles tends to include some apparent outlier fibers. Fig. 4 gives a visualization of three example clusters and their connected FreeSurfer regions. The clusters from DFC are more anatomically coherent, connecting to the same cortical regions.

In addition to the visualization of clusters from individual subjects, we also provide a visual comparison of population-wise clusters to demonstrate the methods' performance for tractography atlas creation. To do so, we compare the DFC and WhiteMatterAnalysis methods, which are explicitly designed to perform groupwise clustering to create tractography atlases. For DFC, the population-wise atlas is derived from our training process, where fiber clusters from the training subjects are formed. For WhiteMatterAnalysis, we use the anatomically curated white matter atlas created using WhiteMatterAnalysis (Zhang et al., 2018c). Fig. 5 gives a visual comparison of results from DFC and WhiteMatterAnalysis. Example clusters are shown in regions of the arcuate fasciculus, corpus callosum and superficial fronto-parietal tracts. It can be seen that the DFC method obtains population-wise clusters that are more separated and compact, where cluster subdivisions better respect terminating anatomical regions.

We also compared the execution time and memory usage of each comparison algorithm during inference for various data sizes. This experiment was performed on a computer equipped with a 2.1 GHz Intel Xeon E5 CPU (8 DIMMs; 264 GB Memory) and an NVIDIA RTX 1080Ti GPU. Testing datasets were obtained by downsampling densely seeded tractography from one example HCP subject to produce datasets of 250,000, 500,000, 750,000, and 1000,000 fibers (streamlines). As shown in Table 3, it is apparent that both execution time and memory usage increase with increasing data size. For all data sizes, DFC and DFC$_{conf}$ are the most efficient due to the use of GPU computation. QuickBundles is also computationally efficient. DFC, DFC$_{conf}$, and QuickBundles show comparably low memory usage. WhiteMatterAnalysis shows a much longer execution time and larger memory usage than other comparison methods, e.g., 55 GB for 1000,000 fibers, due to the expensive pairwise fiber similarity computation between the subject and atlas tractography data. These results in general demonstrate the high efficiency and low computation cost of the proposed DFC method.

### 3.3.2. Comparison to deep convolutional embedded clustering baseline

We compare the proposed DFC method with the DCEC baseline method, which is a widely used auto-encoder model for unsupervised clustering in computer vision (Guo et al., 2017). The inputs of DCEC are expected to be images, and thus we used FiberMap (Zhang et al., 2020) to represent input fibers as images (Chen et al., 2021). Hyperparameters in DCEC were optimized to obtain the best performance.

As shown in Table 4, DFC has obviously improved performance in terms of DB index, TAPC and TSPC, while DCEC has a slightly higher WMPG score (attributed to the lack of outlier removal in DCEC). It is worth noting that, for the DB index, the baseline DCEC obtained an exceptionally large score due to its sensitivity to point order along fibers. Fig. 6 gives a visualization of example clusters from DFC and DCEC, colored by the sequence of points along fibers. We can observe that DFC can successfully group spatially close fibers with opposite point orders into one cluster while DCEC failed to do that.

### 3.3.3. Ablation study

We performed an ablation study to investigate how different modules in the proposed DFC method influence white matter fiber clustering performance. Evaluation of four models was performed, including DFC$_{no-roi\&cor\&ro}$ (DFC without anatomical region, cortical parcellation or outlier removal), DFC_no-cor&ro (DFC without cortical parcellation or outlier removal but with anatomical region), DFC_no-ro (DFC without outlier removal but with anatomical region and cortical parcellation) and our proposed DFC method.

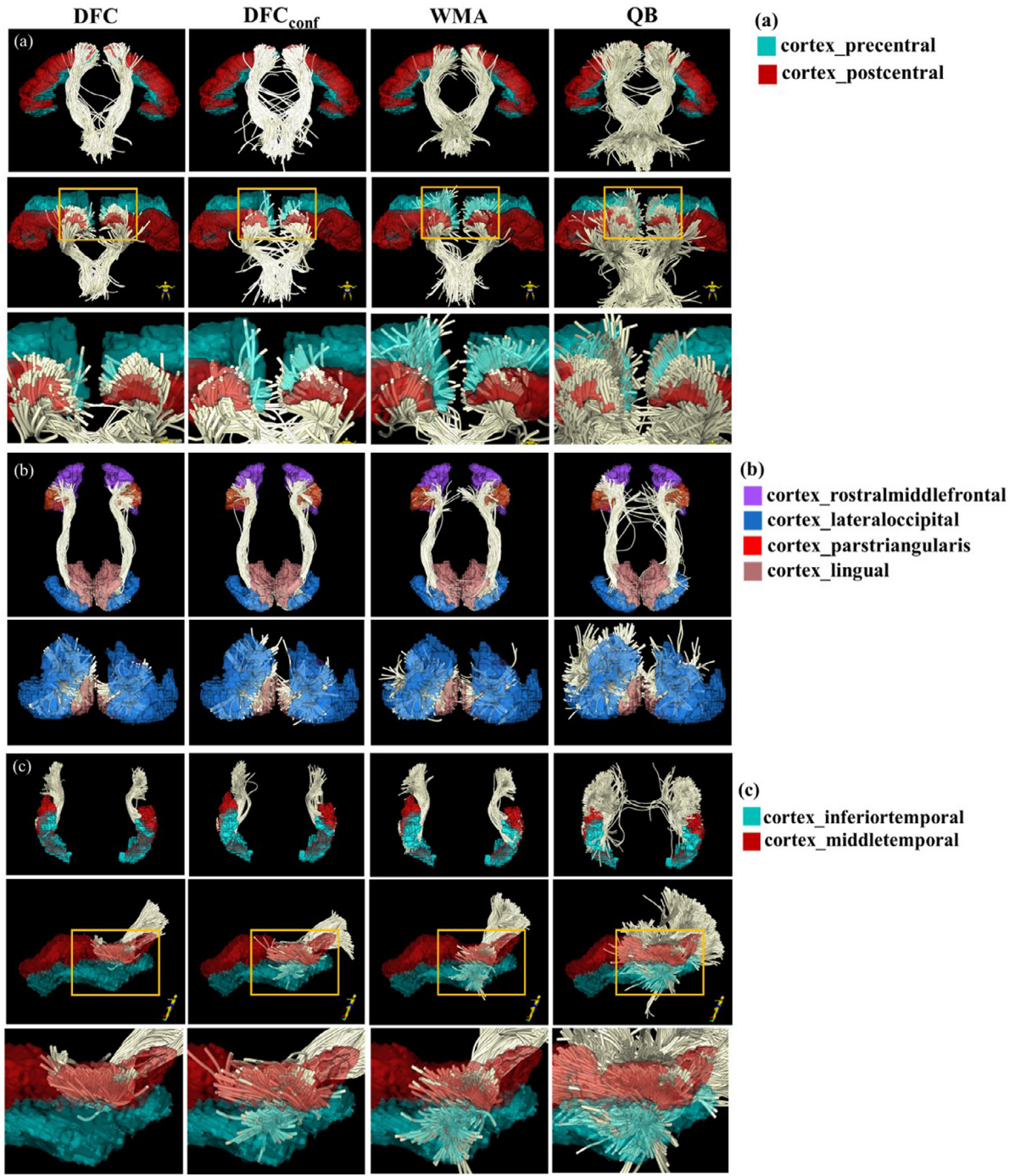

**Fig. 4.** Example clusters for visualization of coherence between clusters and cortical parcels, across different methods (DFC, DFC_conf, WMA and QB) from HCP data. Clusters within the CST, IoFF and Sup_PT tracts are shown in (a), (b) and (c) respectively. For (a) and (c), the first row displays a posterior view; In the second row, the display view is indicated by the human figure at the bottom right corner; The third row is a zoomed-in area of the orange rectangle area in the second row. In (b), the first and second rows show the inferior and posterior view of the IoFF cluster. (Abbreviations: CST - corticospinal tract; IoFF - inferior occipito-frontal fasciculus; Sup_PT - superficial parieto-temporal; DFC - Deep Fiber Clustering; DFC_conf - conference version of Deep Fiber Clustering; WMA - WhiteMatterAnalysis; QB - QuickBundles).

As shown in Table 5, adding anatomical region information improved TAPC (DFC_no-cor&ro vs. DFC_no-roi&cor&ro), adding cortical parcellation information improved TSPC (DFC_no-ro vs. DFC_no-cor&ro), and performing outlier removal improved DB index (DFC_no-ro vs. DFC). The proposed method included these three modules and achieved the best DB, TAPC, and TSPC results. However, we noticed a slight decrease of WMPG (0.3%) due to the removal of false positive fibers during the outlier removal process.

### 3.3.4. Comparison of input representations

We compared three kinds of representations for tractography fibers, i.e., FiberMap, graph and point cloud. For each representation, neural networks that can effectively process the input were used: Convolutional Neural Networks (CNNs) for FiberMap, Graph Convolutional Networks (GCNs) for graphs, and DGCNNs (proposed) for point clouds. The FiberMap input was introduced in Section 3.3.1, and more details can be found in (Zhang et al., 2019a). For the graph input, a fiber (streamline) was naturally regarded as a graph, with points as nodes and edges constructed between adjacent points, analogous to traditional graph construction for meshes in computer vision (Pfaff et al., 2021). The point cloud input was described in Section 2.1. For each input representation and its network, the proposed self-supervised learning pipeline was applied to generate clusters, followed by the proposed outlier removal process. For a fair comparison, we adjusted the threshold in each method so that they removed approximately the same number of fibers.

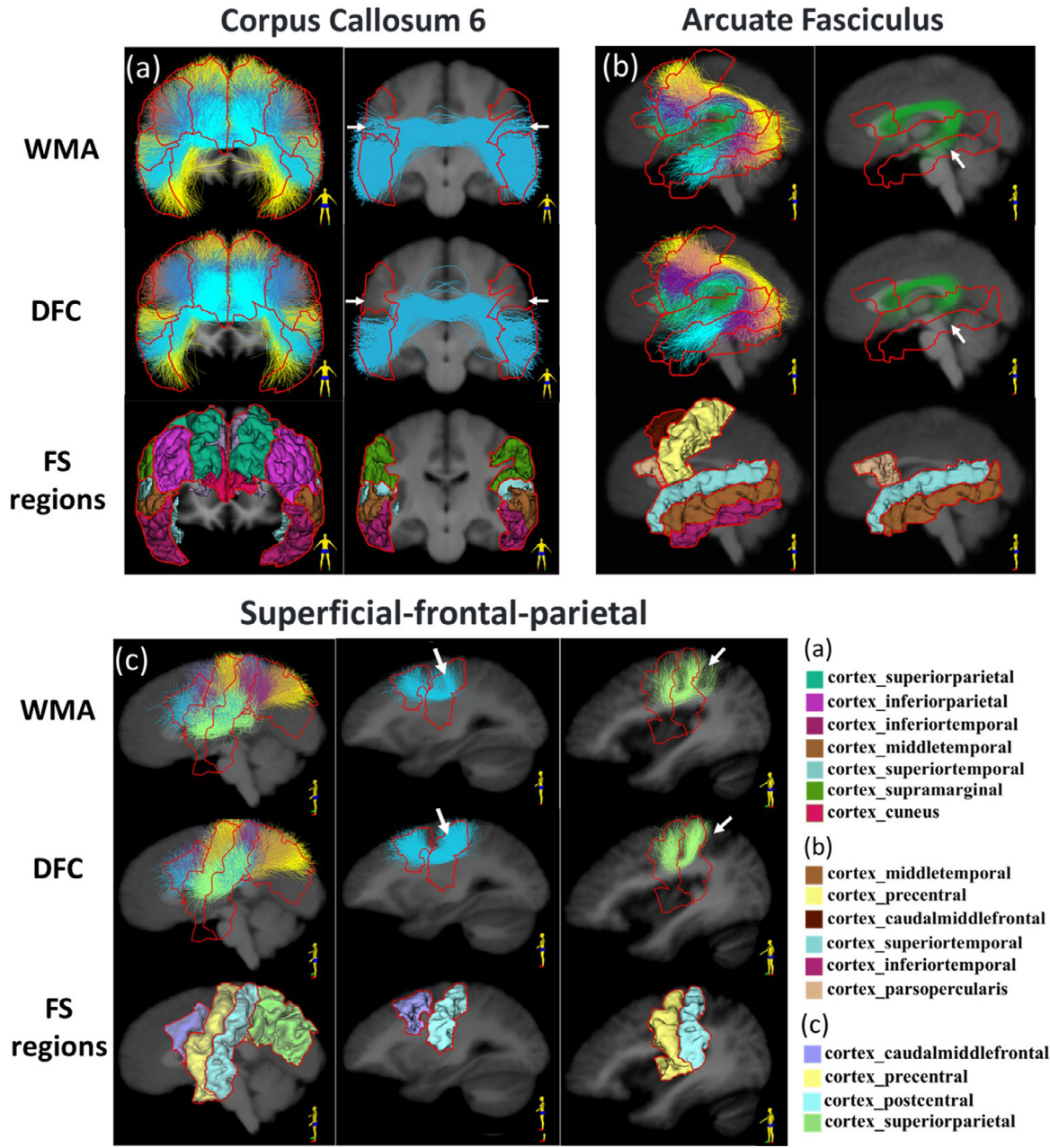

**Fig. 5.** Example tracts for visualization of cluster subdivisions within a tract, across DFC and WMA methods from HCP data. Part of the AF, CC6 and Sup_FP tracts with one or two example clusters are shown in (a), (b) and (c) respectively. The display views are indicated by the human figure at the bottom right corner. (Abbreviations: AF - arcuate fasciculus; CC6 - corpus callosum 6; Sup_FP - superficial fronto-parietal; DFC - Deep Fiber Clustering; WMA - WhiteMatterAnalysis).

**Table 3**
Execution time and memory usage of comparison methods for various data sizes. (Abbreviations: DFC - Deep Fiber Clustering; DFC_conf - conference version of Deep Fiber Clustering; WMA - WhiteMatterAnalysis; QB - QuickBundles).

| data size (number of fibers) | DFC | | DFC_conf | | WMA | | QB | |
|---|---|---|---|---|---|---|---|---|
| | time (sec) | memory usage (GB) | time (sec) | memory usage (GB) | time (sec) | memory usage (GB) | time (sec) | memory usage (GB) |
| 250,000 | 10 | 0.615 | 18 | 0.447 | 2196 | 14.115 | 33 | 0.411 |
| 500,000 | 20 | 1.206 | 36 | 0.893 | 3373 | 28.091 | 61 | 0.822 |
| 750,000 | 31 | 1.795 | 53 | 1.339 | 4582 | 42.066 | 97 | 1.240 |
| 1000,000 | 40 | 2.385 | 74 | 1.786 | 5758 | 54.939 | 117 | 1.640 |

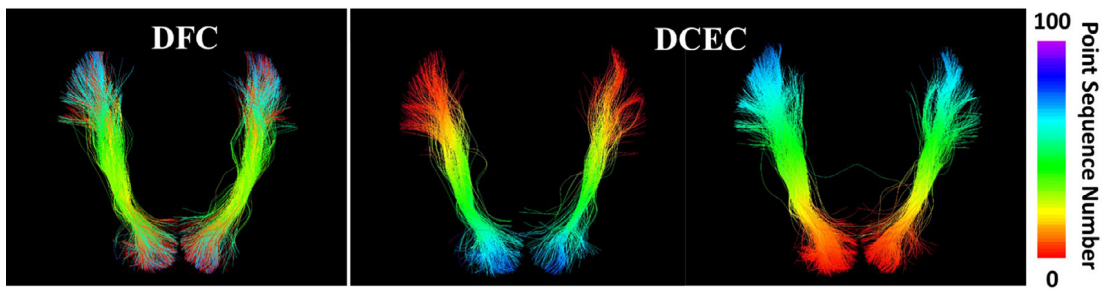

**Fig. 6.** Visualization of example corresponding clusters from DFC and DCEC. Colors represent the sequence number of points along the fiber (rainbow coloring with red for starting and purple for ending points).

**Table 4**
Comparison with DCEC baseline method. Results are presented as mean value with standard deviation across subjects in parenthesis.

|      | Method | DB            | WMPG          | TAPC          | TSPC          |
|------|--------|---------------|---------------|---------------|---------------|
| HCP  | DFC    | **2.014 (0.023)** | 0.996 (0.004) | **0.844 (0.003)** | **0.601(0.008)** |
|      | DCEC   | 15.36 (1.708) | **0.999 (0.003)** | 0.768 (0.004) | 0.459(0.008)  |
| CNP  | DFC    | **2.127 (0.027)** | 0.970 (0.015) | **0.830 (0.002)** | **0.498(0.007)** |
|      | DCEC   | 14.25 (1.660) | **0.994 (0.006)** | 0.745 (0.004) | 0.353(0.009)  |
| PPMI | DFC    | **2.119 (0.028)** | 0.978 (0.014) | **0.832(0.003)** | **0.476(0.009)** |
|      | DCEC   | 14.76 (3.157) | **0.997 (0.004)** | 0.745(0.005)  | 0.329(0.010)  |

**Table 5**
Ablation Study for DFC. Results are presented as mean value with standard deviation across subjects in parenthesis.

|      | DFC$_{no\text{-}roi\&cor\&ro}$ | DFC$_{no\text{-}cor\&ro}$ | DFC$_{no\text{-}ro}$ | DFC           |
|------|--------------------------------|---------------------------|----------------------|---------------|
| DB   | 2.278 (0.029)                  | 2.292 (0.031)             | 2.336 (0.034)        | **2.014 (0.023)** |
| WMPG | **0.999 (0.002)**              | **0.999 (0.002)**         | **0.999 (0.002)**    | 0.996 (0.004) |
| TAPC | 0.792 (0.004)                  | 0.811 (0.003)             | 0.814 (0.003)        | **0.844 (0.003)** |
| TSPC | 0.495 (0.008)                  | 0.508 (0.008)             | 0.537 (0.008)        | **0.601 (0.008)** |

**Table 6**
Comparison of different input representations and corresponding neural networks. Results are presented as the mean value with the standard deviation across subjects in parenthesis.

|      | DGCNN + point cloud | GCN + graph   | CNN + FiberMap |
|------|---------------------|---------------|----------------|
| DB   | 2.014 (0.023)       | **2.006 (0.018)** | 2.017 (0.022)  |
| WMPG | 0.996 (0.004)       | **0.997 (0.005)** | **0.997 (0.004)** |
| TAPC | **0.844 (0.003)**   | 0.840 (0.003) | 0.842 (0.002)  |
| TSPC | **0.601 (0.008)**   | 0.593 (0.007) | 0.597 (0.007)  |

As shown in Table 6, the three models with different input representations all demonstrate good performance in terms of the four evaluation metrics, indicating the effectiveness of our network design. The DGCNN model with a point cloud representation shows the best performance in general, with the shortest execution time (~15 s) on one randomly selected HCP subject (about 400k fibers). Though GCN with a graph representation has the lowest DB index, its TAPC and TSPC scores are lower than DGCNN, and its prediction time is much longer (~110 s) than the others. Compared to the CNN with a FiberMap representation, DGCNN has better performance in the DB index, TAPC and TSPC scores as well as the computation time (~30 s for CNN), though the WMPG is slightly lower.

### 3.3.5. Comparison of outlier removal methods

We provide a visual comparison between two outlier removal strategies: RO$_{absolute}$ that adopts an absolute removal threshold for all clusters (proposed in our conference paper version), and RO$_{adaptive}$ that adopts a cluster-adaptive threshold (proposed in the present work). For RO$_{absolute}$, the threshold was set to 0.045 so that it removed a similar percentage of fibers as RO$_{adaptive}$ (0.2626 and 0.2571, respectively).

As shown in Fig. 7, the results of RO$_{adaptive}$ are more anatomically plausible, while the compared RO$_{absolute}$ method tends to be overly strict (Fig. 7a) or not properly reject apparent outlier fibers (Fig. 7b).

## 4. Discussion

In this work, we proposed a novel end-to-end unsupervised deep learning framework, DFC, for fast and effective white matter fiber clustering. Our clustering method leverages not only white matter fiber geometry information but also gray matter anatomical parcellation information. The performance of DFC was evaluated on three independently acquired datasets across genders, ages and health conditions. Several detailed observations about the experimental results are discussed below.

Our method demonstrated advanced performance compared to several state-of-the-art methods in terms of cluster compactness, anatomical coherence, generalization ability and efficiency. WhiteMatterAnalysis has demonstrated consistent white matter fiber clustering across independently acquired datasets from different populations (Zhang et al., 2018c). Our results in the three testing datasets support this finding regarding the generalization of WhiteMatterAnalysis. However, the computational time of WhiteMatterAnalysis is much longer compared to the other state-of-the-art methods (QuickBundles, DFC and DFC$_{conf}$). QuickBundles generated compact and well-separated clusters within each subject, but it was not designed to generalize to a population. Compared to DFC$_{conf}$, the cortical anatomical coherence of clusters from DFC was improved due to the incorporation of cortical parcellation information. However, DFC showed slightly decreased TAPC compared to DFC$_{conf}$, likely because the incorporation of cortical surface parcellation information reduces the contribution of anatomical region information in cluster assignment. Unlike some unsupervised clustering methods like QuickBundles, our method requires a training stage, which takes about three hours. However, the training process only needs to be performed once and the model trained on one dataset can be generalized to other datasets. In addition, the inference time of DFC is very low, making it an efficient fiber clustering method.

Our pipeline adopts the self-supervised learning strategy to learn deep embeddings for unsupervised fiber clustering. Many pretext tasks,

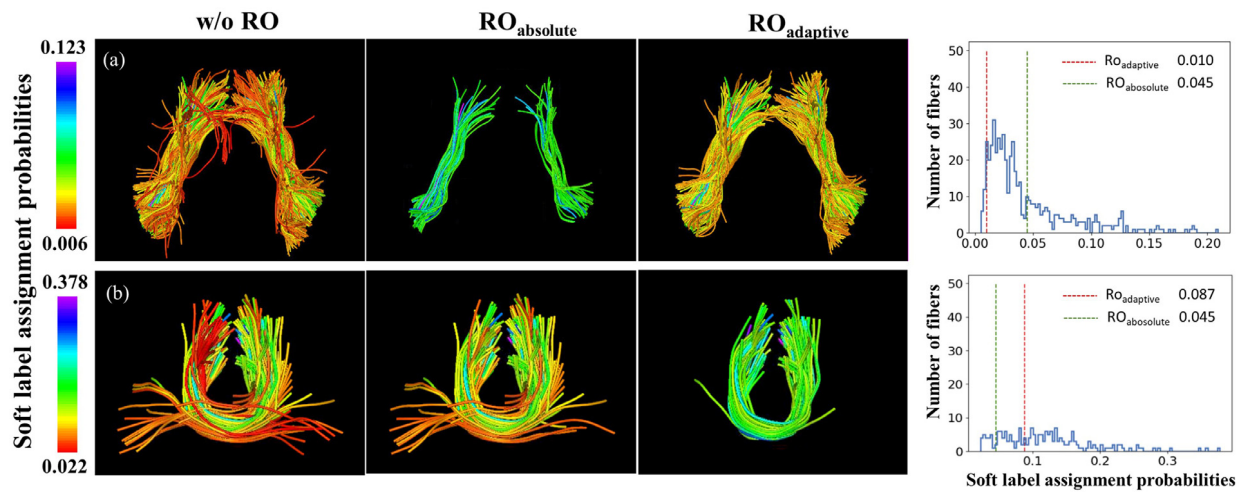

**Fig. 7.** Example clusters to compare previous (RO$_{absolute}$) and current outlier removal methods (RO$_{adaptive}$). Results of two clusters (two rows) from no outlier removal (w/o RO), RO$_{absolute}$ and RO$_{adaptive}$ methods are displayed in columns 1-3 respectively. The fiber color indicates the soft label assignment probability of the fiber (rainbow coloring with red indicating the smallest and purple the largest). The fourth column shows the soft label assignment probability distribution within the selected clusters. The red and green dashed lines indicate the thresholds of outlier removal for RO$_{adaptive}$ and RO$_{absolute}$, respectively.

such as predicting context (Doersch et al., 2015) or image rotation (Komodakis and Gidaris, 2018), have been proposed in the computer vision community (Chen et al., 2020; Liu et al., 2021; Zhang et al., 2016). For medical image computing tasks, novel pretext tasks are designed by harnessing knowledge from the medical domain instead of directly adopting pre-designed pretext tasks from the computer vision field (Matzkin et al., 2020; Shurrab and Duwairi, 2022; Spitzer et al., 2018; P. Zhang et al., 2017). In our DFC framework, we designed the pretext task of fiber distance prediction to obtain embeddings for subsequent clustering. The minimum average direct-flip distance adopted in our study can be easily replaced with other fiber distance measures of interest such as the mean closest point fiber distance (L. J. O'Donnell and Westin, 2007) or Hausdorff distance (Corouge et al., 2004b). The pretext task leverages domain-specific knowledge of fiber distance, which can provide the following advantages. First, the general idea of white matter fiber clustering is to group fibers with low pairwise distances into the same group. By solving the pretext task of fiber distance prediction, our pipeline obtains embeddings whose pairwise distances are consistent with their corresponding fibers and thus benefits the performance of white matter fiber clustering. Second, the proposed self-supervised learning strategy could guide the network to learn similar embeddings for spatially close fibers regardless of their fiber point orderings, enabling them to be grouped into the same cluster. This gives our method an advantage over the widely used auto-encoder based models (Guo et al., 2017; Xie et al., 2016), which are sensitive to fiber point ordering because they learn embeddings by reconstructing the input itself.

We proposed a novel framework that enables combined use of white matter fiber geometry and gray matter anatomical parcellation information in white matter fiber clustering. Most current white matter fiber clustering methods group fibers into bundles by calculating the similarity of fibers based on their coordinates in Euclidean space (Garyfallidis et al., 2012; Vázquez et al., 2020b). On the other hand, a recent study performed white matter fiber clustering based on the brain anatomical structures each fiber passes through instead of fiber spatial coordinates (Siless et al., 2018). Therefore, either source of information could make contributions to the white matter fiber clustering task. In our method, we perform clustering leveraging both sources of information, including the spatial coordinates of fibers and gray matter anatomical parcellation information, to help identify anatomically meaningful clusters. The results show that integrating gray matter anatomical parcellation information clearly improved the anatomical coherence within clusters. Therefore, anatomical parcellation information provides useful complementary information to fiber geometric information. However, we only investigated the performance of the Desikan-Killiany parcellation (Desikan et al., 2006). A finer parcellation that provides more detailed information, such as that defined in more recently proposed atlases (Destrieux et al., 2010; Glasser et al., 2016), may be more beneficial to clustering performance. Our method shows the potential of combining multiple sources of information to improve white matter fiber clustering.

The representation of tractography data for deep learning is an open challenge for tractography-related tasks. Previous studies performed tractography segmentation by working on 3D volumes instead of the tractography data (Liu et al., 2022; Lu et al., 2020; Wasserthal et al., 2018), but this neglects subject-specific fiber tractography information. Recently, FiberMap was proposed to represent a fiber as a 2D image (Zhang et al., 2020, 2019a), a sparse representation of fibers that needs an extra step to generate. In our work, we used point clouds to represent fibers. Point clouds are compact representations of the original fiber points and enable end-to-end learning of the neural network. In addition, point-based models are permutation invariant to input points and thus insensitive to point ordering along fibers. By representing fibers as point clouds, we adopted point-based neural networks, which show good clustering performance as well as efficiency.

In this study, we propose a simple but effective outlier removal strategy to filter anatomically implausible fibers and improve white matter fiber clustering performance. Our strategy is rapid, as it simply rejects outlier fibers with low cluster assignment probabilities, without any added computational burden of fiber distance computations (Zhang et al., 2018c) or convex optimization (Daducci et al., 2015). However, our simple strategy is only able to remove fibers that do not correspond well to a cluster. We expect that a combination of outlier removal methods may have the best performance for reducing the well-known impact of outliers on fiber tractography (Drakesmith et al., 2015).

Limitations and potential future directions of the current work are as follows. First, our proposed pipeline only combines two sources of information, i.e., white matter fiber geometry and gray matter anatomical parcellation information, to achieve white matter fiber clustering. It is worth investigating incorporating additional sources of information such as functional MRI to obtain functionally meaningful clusters. Future work could also investigate more advanced neural networks and other self-supervised learning strategies such as contrastive learning (Chen et al., 2020) to potentially obtain better clustering results.

## 5. Conclusion

In this paper, we present a novel end-to-end unsupervised deep learning framework for white matter fiber clustering. We adopt the self-supervised learning strategy to enable joint deep embedding and cluster assignment. Our method can handle several key challenges in white matter fiber clustering methods including improving implementation efficiency, handling flipped order of points along fibers, combining fiber geometric and anatomical information, filtering anatomically implausible fibers and inter-subject correspondence of fiber clusters. Experimental results show that our proposed method achieves fast and effective white matter fiber clustering and demonstrates advantages over state-of-the-art algorithms in terms of clustering performance as well as efficiency.

## Declaration of Competing Interest

None.

## Credit authorship contribution statement

**Yuqian Chen:** Conceptualization, Methodology, Software, Writing – original draft. **Chaoyi Zhang:** Methodology, Writing – review & editing. **Tengfei Xue:** Methodology, Writing – review & editing. **Yang Song:** Writing – review & editing. **Nikos Makris:** Conceptualization, Writing – review & editing. **Yogesh Rathi:** Writing – review & editing. **Weidong Cai:** Resources, Supervision, Writing – review & editing. **Fan Zhang:** Conceptualization, Methodology, Writing – review & editing. **Lauren J. O'Donnell:** Conceptualization, Methodology, Writing – review & editing, Supervision, Funding acquisition.

## Data and code availability

The data used in this project is from three datasets, Human Connectome Project (HCP), Consortium for Neuropsychiatric Phenomics (CNP) and Parkinson Progression Marker Initiative (PPMI). The three datasets can be downloaded through ConnectomeDB (db.humanconnectome.org), OpenFMRI project (http://openfmri.org) and PPMI website (www.ppmi-info.org) respectively. The data analysis code is available on https://github.com/SlicerDMRI/DFC.

## Acknowledgments

We acknowledge funding provided by the following National Institutes of Health (NIH) grants: R01MH125860, R01MH119222, R01MH074794, R01NS125781 and P41EB015902. F.Z. also acknowledges a BWH Radiology Research Pilot Grant Award.

## Supplementary materials

Supplementary material associated with this article can be found, in the online version, at doi:10.1016/j.neuroimage.2023.120086.

## References

Astolfi, P., Verhagen, R., Petit, L., Olivetti, E., Masci, J., Boscaini, D., Avesani, P., 2020. Tractogram Filtering of Anatomically Non-Plausible Fibers with Geometric Deep Learning, in: Medical Image Computing and Computer Assisted Intervention – MICCAI 2020. Springer International Publishing, pp. 291–301.

Avants, B.B., Tustison, N., Song, G., 2009. Advanced normalization tools (ANTS). Insight J. 2, 1–35.

Basser, P.J., Pajevic, S., Pierpaoli, C., Duda, J., Aldroubi, A., 2000. In vivo fiber tractography using DT-MRI data. Magn. Reson. Med. 44, 625–632.

Battocchio, M., Schiavi, S., Descoteaux, M., Daducci, A., 2022. Bundle-o-graphy: improving structural connectivity estimation with adaptive microstructure-informed tractography. Neuroimage 263, 119600.

Brun, A., Knutsson, H., Park, H.-J., Shenton, M.E., Westin, C.-F., 2004. Clustering Fiber Traces Using Normalized Cuts, in: Medical Image Computing and Computer Assisted Intervention – MICCAI 2004. Springer International Publishing, pp. 368–375.

Chekir, A., Descoteaux, M., Garyfallidis, E., Côté, M., Boumghar, F.O., 2014. A hybrid approach for optimal automatic segmentation of White Matter tracts in HARDI. In: 2014 IEEE Conference on Biomedical Engineering and Sciences (IECBES). IEEE, pp. 177–180.

Chen, T., Kornblith, S., Norouzi, M., Hinton, G., 2020. A simple framework for contrastive learning of visual representations. In: Iii, H.D., Singh, A. (Eds.), Proceedings of the 37th International Conference on Machine Learning, Proceedings of Machine Learning Research. PMLR, pp. 1597–1607.

Chen, X., Ma, H., Wan, J., Li, B., Xia, T., 2017. Multi-view 3d object detection network for autonomous driving. In Proceedings of the IEEE Conference on Computer Vision and Pattern Recognition, pp. 1907–1915.

Chen, Y., Zhang, C., Song, Y., Makris, N., Rathi, Y., Cai, W., Zhang, F., O'Donnell, L.J., 2021. Deep Fiber Clustering: Anatomically Informed Unsupervised Deep Learning for Fast and Effective White Matter Parcellation, in: Medical Image Computing and Computer Assisted Intervention – MICCAI 2021. Springer International Publishing, pp. 497–507.

Chen, Y., Zhang, F., Zekelman, L.R., Xue, T., Zhang, C., Song, Y., Makris, N., Rathi, Y., Cai, W., O'Donnell, L.J., 2023. TractGraphCNN: anatomically informed graph CNN for classification using diffusion MRI tractography. 2023 IEEE 20th International Symposium on Biomedical Imaging (ISBI) (accepted). IEEE.

Chen, Y., Zhang, F., Zhang, C., Xue, T., Zekelman, L.R., He, J., Song, Y., Makris, N., Rathi, Y., Golby, A.J., Cai, W., O'Donnell, L.J., 2022. White matter tracts are point clouds: neuropsychological score prediction and critical region localization via geometric deep learning. In: Medical Image Computing and Computer Assisted Intervention – MICCAI 2022. Springer Nature Switzerland, pp. 174–184.

Chopra, S., Hadsell, R., LeCun, Y., 2005. Learning a similarity metric discriminatively, with application to face verification. In: IEEE Computer Society Conference on Computer Vision and Pattern Recognition (CVPR'05), pp. 539–546.

Ciccarelli, O., Catani, M., Johansen-Berg, H., Clark, C., Thompson, A., 2008. Diffusion-based tractography in neurological disorders: concepts, applications, and future developments. Lancet Neurol. 7, 715–727.

Corouge, I., Gouttard, S., Gerig, G., 2004a. Towards a shape model of white matter fiber bundles using diffusion tensor MRI. In: 2004 2nd IEEE International Symposium on Biomedical Imaging: Nano to Macro, 1, pp. 344–347.

Corouge, I., Gouttard, S., Gerig, G., 2004b. Towards a shape model of white matter fiber bundles using diffusion tensor MRI. In: 2004 2nd IEEE International Symposium on Biomedical Imaging: Nano to Macro (IEEE Cat No. 04EX821), 1, pp. 344–347.

Cousineau, M., Jodoin, P.-M., Morency, F.C., Rozanski, V., Grand'Maison, M., Bedell, B.J., Descoteaux, M., 2017. A test-retest study on Parkinson's PPMI dataset yields statistically significant white matter fascicles. Neuroimage Clin. 16, 222–233.

Daducci, A., Dal Palù, A., Lemkaddem, A., Thiran, J.-P., 2015. COMMIT: convex optimization modeling for microstructure informed tractography. IEEE Trans. Med. Imaging 34, 246–257.

Dave, D., Varma, T., 2014. A review of various statistical methods for outlier detection. Int. J. Comput. Sci. Eng. Technol. (IJCSET) 5, 137–140.

Desikan, R.S., Ségonne, F., Fischl, B., Quinn, B.T., Dickerson, B.C., Blacker, D., Buckner, R.L., Dale, A.M., Maguire, R.P., Hyman, B.T., Albert, M.S., Killiany, R.J., 2006. An automated labeling system for subdividing the human cerebral cortex on MRI scans into gyral based regions of interest. Neuroimage 31, 968–980.

Destrieux, C., Fischl, B., Dale, A., Halgren, E., 2010. Automatic parcellation of human cortical gyri and sulci using standard anatomical nomenclature. Neuroimage 53, 1–15.

Doersch, C., Gupta, A., Efros, A.A., 2015. Unsupervised visual representation learning by context prediction. In: Proceedings of the IEEE International Conference on Computer Vision, pp. 1422–1430.

Drakesmith, M., Caeyenberghs, K., Dutt, A., Lewis, G., David, A.S., Jones, D.K., 2015. Overcoming the effects of false positives and threshold bias in graph theoretical analyses of neuroimaging data. Neuroimage 118, 313–333.

Essayed, W.I., Zhang, F., Unadkat, P., Cosgrove, G.R., Golby, A.J., O'Donnell, L.J., 2017. White matter tractography for neurosurgical planning: a topography-based review of the current state of the art. Neuroimage Clin. 15, 659–672.

Fischl, B., 2012. FreeSurfer. Neuroimage 62, 774–781.

Garyfallidis, E., Brett, M., Correia, M.M., Williams, G.B., Nimmo-Smith, I., 2012. Quick-Bundles, a method for tractography simplification. Front. Neurosci. 6, 175.

Garyfallidis, E., Côté, M.-A., Rheault, F., Descoteaux, M., 2016. QuickBundlesX: sequential clustering of millions of streamlines in multiple levels of detail at record execution time. 24th International Society of Magnetic Resonance in Medicine (ISMRM) (2016).

Garyfallidis, E., Côté, M.-A., Rheault, F., Sidhu, J., Hau, J., Petit, L., Fortin, D., Cunanne, S., Descoteaux, M., 2018. Recognition of white matter bundles using local and global streamline-based registration and clustering. Neuroimage 170, 283–295.

Glasser, M.F., Coalson, T.S., Robinson, E.C., Hacker, C.D., Harwell, J., Yacoub, E., Ugurbil, K., Andersson, J., Beckmann, C.F., Jenkinson, M., Smith, S.M., Van Essen, D.C., 2016. A multi-modal parcellation of human cerebral cortex. Nature 536, 171–178.

Gong, G., He, Y., Concha, L., Lebel, C., Gross, D.W., Evans, A.C., Beaulieu, C., 2009. Mapping anatomical connectivity patterns of human cerebral cortex using in vivo diffusion tensor imaging tractography. Cereb. Cortex 19, 524–536.

Guevara, M., Guevara, P., Román, C., Mangin, J.-F., 2020. Superficial white matter: a review on the dMRI analysis methods and applications. Neuroimage 212, 116673.

Guevara, M., Sun, Z.-Y., Guevara, P., Rivière, D., Grigis, A., Poupon, C., Mangin, J.-F., 2022. Disentangling the variability of the superficial white matter organization using regional-tractogram-based population stratification. Neuroimage 255, 119197.

Guevara, P., Duclap, D., Poupon, C., Marrakchi-Kacem, L., Fillard, P., Le Bihan, D., Leboyer, M., Houenou, J., Mangin, J.-F., 2012. Automatic fiber bundle segmenta-tion in massive tractography datasets using a multi-subject bundle atlas. Neuroimage 61, 1083–1099.

Guevara, P., Poupon, C., Rivière, D., Cointepas, Y., Descoteaux, M., Thirion, B., Mangin, J.-F., 2011. Robust clustering of massive tractography datasets. Neuroimage 54, 1975–1993.

Guo, X., Liu, X., Zhu, E., Yin, J., 2017. Deep clustering with convolutional autoencoders. In: International Conference on Neural Information Processing, pp. 373–382.

Gupta, T., Patil, S.M., Tailor, M., Thapar, D., Nigam, A., 2017. BrainSegNet: a segmentation network for human brain fiber tractography data into anatomically meaningful clusters. arXiv [cs.CV]: 1710.05158.

Gupta, V., Thomopoulos, S.I., Rashid, F.M., 2017b. FiberNET: An ensemble Deep Learning Framework For Clustering White Matter fibers, in: Medical Image Computing and Computer Assisted Intervention – MICCAI 2017. Springer International Publishing, pp. 548–555.

Hassani, K., Haley, M., 2019. Unsupervised multi-task feature learning on point clouds. In: Proceedings of the IEEE/CVF International Conference on Computer Vision, pp. 8160–8171.

He, K., Gkioxari, G., Dollár, P., Girshick, R., 2017. Mask r-cnn. In: Proceedings of the IEEE International Conference on Computer Vision, pp. 2961–2969.

Huerta, I., Vazquez, A., Lopez-Lopez, N., Houenou, J., Poupon, C., Mangin, J.-F., Guevara, P., Hernandez, C., 2020. Inter-subject clustering of brain fibers from whole-brain tractography. Conf. Proc. IEEE Eng. Med. Biol. Soc. 2020, 1687–1691.

Ji, E., Guevara, P., Guevara, M., Grigis, A., Labra, N., Sarrazin, S., Hamdani, N., Bellivier, F., Delavest, M., Leboyer, M., Tamouza, R., Poupon, C., Mangin, J.-F., Houenou, J., 2019. Increased and decreased superficial white matter structural connectivity in schizophrenia and bipolar disorder. Schizophr. Bull. 45, 1367–1378.

Jin, Y., Shi, Y., Zhan, L., Gutman, B.A., de Zubicaray, G.I., McMahon, K.L., Wright, M.J., Toga, A.W., Thompson, P.M., 2014. Automatic clustering of white matter fibers in brain diffusion MRI with an application to genetics. Neuroimage 100, 75–90.

Károly, A.I., Fullér, R., Galambos, P., 2018. Unsupervised clustering for deep learning: a tutorial survey. Acta Polytech. Hungarica 15, 29–53.

Kingma, D.P., Ba, J., 2014. Adam: a method for stochastic optimization. arXiv [cs.LG]: 1412.6980.

Kolesnikov, A., Zhai, X., Beyer, L., 2019. Revisiting self-supervised visual representation learning. In: Proceedings of the IEEE/CVF Conference on Computer Vision and Pattern Recognition, pp. 1920–1929.

Komodakis, Gidaris, 2018. Unsupervised representation learning by predicting image rotations. International Conference on Learning Representations (ICLR).

Legarreta, J.H., Petit, L., Rheault, F., Theaud, G., Lemaire, C., Descoteaux, M., Jodoin, P.-M., 2021. Filtering in tractography using autoencoders (FINTA). Med. Image Anal. 72, 102126.

Levitt, J.J., Zhang, F., Vangel, M., Nestor, P.G., Rathi, Y., Kubicki, M., Shenton, M.E., O'Donnell, L.J., 2021. The organization of frontostriatal brain wiring in healthy subjects using a novel diffusion imaging fiber cluster analysis. Cereb. Cortex 31, 5308–5318.

Li, H., Xue, Z., Guo, L., Liu, T., Hunter, J., Wong, S.T.C., 2010. A hybrid approach to automatic clustering of white matter fibers. Neuroimage 49, 1249–1258.

Likas, A., Vlassis, N., Verbeek, J., 2003. The global k-means clustering algorithm. Pattern Recognit. 36, 451–461.

Liu, F., Feng, J., Chen, G., Wu, Y., Hong, Y., Yap, P.-T., Shen, D., 2019. DeepBundle: fiber bundle parcellation with graph convolution neural networks. In: International Workshop on Graph Learning in Medical Imaging, pp. 88–95.

Liu, W., Lu, Q., Zhuo, Z., Li, Y., Duan, Y., Yu, P., Qu, L., Ye, C., Liu, Y., 2022. Volumetric segmentation of white matter tracts with label embedding. Neuroimage 250, 118934.

Liu, X., Zhang, F., Hou, Z., Mian, L., Wang, Z., Zhang, J., Tang, J., 2021. Self-supervised learning: generative or contrastive. IEEE Trans. Knowl. Data Eng. 1.

Logiraj, K., Thanikasalam, K., Sotheeswaran, S., Ratnarajah, N., 2021. TractNet: a deep learning approach on 3D curves for segmenting white matter fibre bundles. In: 2021 21st International Conference on Advances in ICT for Emerging Regions (ICter), pp. 75–80.

Lu, Q., Li, Y., Ye, C., 2020. White Matter Tract Segmentation with Self-supervised Learning, in: Medical Image Computing and Computer Assisted Intervention – MICCAI 2020. Springer International Publishing, pp. 270–279.

Maaten, L.V.D, 2008. Visualizing datausing t-SNE. J. Mach. Learn. Res. 9, 2579–2605.

Maddah, M., Zöllei, L., Grimson, W.E.L., Westin, C.-F., Wells, W.M., 2008. A mathematical framework for incorporating anatomical knowledge in DT-MRI Analysis. In: 2008 5th IEEE International Symposium on Biomedical Imaging, pp. 105–108.

Maier-Hein, K.H., Neher, P.F., Houde, J.-C., Côté, M.-A., Garyfallidis, E., Zhong, J., Chamberland, M., Yeh, F.-C., Lin, Y.-C., Ji, Q., Reddick, W.E., Glass, J.O., Chen, D.Q., Feng, Y., Gao, C., Wu, Y., Ma, J., He, R., Li, Q., Westin, C.-F., Deslauriers-Gauthier, S., González, J.O.O., Paquette, M., St-Jean, S., Girard, G., Rheault, F., Sidhu, J., Tax, C.M.W., Guo, F., Mesri, H.Y., Dávid, S., Froeling, M., Heemskerk, A.M., Leemans, A., Boré, A., Pinsard, B., Bedetti, C., Desrosiers, M., Brambati, S., Doyon, J., Sarica, A., Vasta, R., Cerasa, A., Quattrone, A., Yeatman, J., Khan, A.R., Hodges, W., Alexander, S., Romascano, D., Barakovic, M., Auría, A., Esteban, O., Lemkaddem, A., Thiran, J.-P., Cetingul, H.E., Odry, B.L., Mailhe, B., Nadar, M.S., Pizzagalli, F., Prasad, G., Villalon-Reina, J.E., Galvis, J., Thompson, P.M., Requejo, F.D.S., Laguna, P.L., Lacerda, L.M., Barrett, R., Dell'Acqua, F., Catani, M., Petit, L., Caruyer, E., Daducci, A., Dyrby, T.B., Holland-Letz, T., Hilgetag, C.C., Stieltjes, B., Descoteaux, M., 2017. The challenge of mapping the human connectome based on diffusion tractography. Nat. Commun. 8, 1349.

Malcolm, J.G., Shenton, M.E., Rathi, Y., 2010. Filtered multitensor tractography. IEEE Trans. Med. Imaging 29, 1664–1675.

Marek, K., Jennings, D., Lasch, S., Siderowf, A., Tanner, C., Simuni, T., Coffey, C., Kieburtz, K., Flagg, E., Chowdhury, S., Poewe, W., Mollenhauer, B., Klinik, P.-E., Sherer, T., Frasier, M., Meunier, C., Rudolph, A., Casaceli, C., Seibyl, J., Men-

dick, S., Schuff, N., Zhang, Y., Toga, A., Crawford, K., Ansbach, A., De Blasio, P., Piovella, M., Trojanowski, J., Shaw, L., Singleton, A., Hawkins, K., Eberling, J., Brooks, D., Russell, D., Leary, L., Factor, S., Sommerfeld, B., Hogarth, P., Pighetti, E., Williams, K., Standaert, D., Guthrie, S., Hauser, R., Delgado, H., Jankovic, J., Hunter, C., Stern, M., Tran, B., Leverenz, J., Baca, M., Frank, S., Thomas, C.-A., Richard, I., Deeley, C., Rees, L., Sprenger, F., Lang, E., Shill, H., Obradov, S., Fernandez, H., Winters, A., Berg, D., Gauss, K., Galasko, D., Fontaine, D., Mari, Z., Gerstenhaber, M., Brooks, D., Malloy, S., Barone, P., Longo, K., Comery, T., Ravina, B., Grachev, I., Gallagher, K., Collins, M., Widnell, K.L., Ostrowizki, S., Fontoura, P., Ho, T., Luthman, J., Brug, M.van der, Reith, A.D., Taylor, P., 2011. The Parkinson progression marker initiative (PPMI). Prog. Neurobiol. 95, 629–635.

Matzkin, F., Newcombe, V., Stevenson, S., Khetani, A., Newman, T., Digby, R., Stevens, A., Glocker, B., Ferrante, E., 2020. Self-supervised Skull Reconstruction in Brain CT Images With Decompressive Craniectomy, in: Medical Image Computing and Computer Assisted Intervention – MICCAI 2020. Springer International Publishing, pp. 390–399.

Mendoza, C., Roman, C., Vazquez, A., Poupon, C., Mangin, J.-F., Hernandez, C., Guevara, P., 2021. Enhanced automatic segmentation for superficial white matter fiber bundles for probabilistic tractography datasets. In: 2021 43rd Annual International Conference of the IEEE Engineering in Medicine & Biology Society (EMBC), pp. 3654–3658.

Mori, S., Crain, B.J., Chacko, V.P., van Zijl, P.C., 1999. Three-dimensional tracking of axonal projections in the brain by magnetic resonance imaging. Ann. Neurol. 45, 265–269.

Ngattai Lam, P.D., Belhomme, G., Ferrall, J., Patterson, B., Styner, M., Prieto, J.C., 2018. TRAFIC: fiber tract classification using deep learning. Med. Imaging 2018: Image Process. 10574, 257–265.

O'Donnell, L.J., Golby, A.J., Westin, C.-F., 2013. Fiber clustering versus the parcellation-based connectome. Neuroimage 80, 283–289.

O'Donnell, L.J., Suter, Y., Rigolo, L., Kahali, P., Zhang, F., Norton, I., Albi, A., Olubiyi, O., Meola, A., Essayed, W.I., Unadkat, P., Ciris, P.A., Wells 3rd, W.M., Rathi, Y., Westin, C.-F., Golby, A.J., 2017. Automated white matter fiber tract identification in patients with brain tumors. Neuroimage Clin. 13, 138–153.

O'Donnell, L.J., Wells 3rd, W.M., Golby, A.J., Westin, C.-F., 2012. Unbiased groupwise registration of white matter tractography. In: Medical Image Computing and Computer Assisted Intervention – MICCAI 2012. Springer International Publishing, pp. 123–130.

O'Donnell, L.J., Westin, C.-F., 2007a. Automatic tractography segmentation using a high-dimensional white matter atlas. IEEE Trans. Med. Imaging 26, 1562–1575.

O'Donnell, L.J., Westin, C.-F., 2007b. Automatic tractography segmentation using a high-dimensional white matter atlas. IEEE Trans. Med. Imaging doi:10.1109/tmi.2007.906785.

O'Donnell, L.J., Westin, C.-F., 2005. White matter tract clustering and correspondence in populations. In: Medical Image Computing and Computer Assisted Intervention – MICCAI 2005. Springer International Publishing, pp. 140–147.

Paszke, A., Gross, S., Massa, F., Lerer, A., Bradbury, J., Chanan, G., Killeen, T., Lin, Z., Gimelshein, N., Antiga, L., Desmaison, A., Kopf, A., Yang, E., DeVito, Z., Raison, M., Tejani, A., Chilamkurthy, S., Steiner, B., Fang, L., Bai, J., Chintala, S., 2019. PyTorch: an imperative style, high-performance deep learning library. Advances in Neural Information Processing Systems 32. Curran Associates, Inc.

Pfaff, T., Fortunato, M., Sanchez-Gonzalez, A., Battaglia, P.W., 2021. Learning mesh-based simulation with graph networks. 2021 9th International Conference on Learning Representations (ICLR).

Piper, R.J., Yoong, M.M., Kandasamy, J., Chin, R.F., 2014. Application of diffusion tensor imaging and tractography of the optic radiation in anterior temporal lobe resection for epilepsy: a systematic review. Clin. Neurol. Neurosurg. 124, 59–65.

Poldrack, R.A., Congdon, E., Triplett, W., Gorgolewski, K.J., Karlsgodt, K.H., Mumford, J.A., Sabb, F.W., Freimer, N.B., London, E.D., Cannon, T.D., Bilder, R.M., 2016. A phenome-wide examination of neural and cognitive function. Sci. Data 3, 160110.

Prasad, G., Joshi, S.H., Jahanshad, N., Villalon-Reina, J., Aganj, I., Lenglet, C., Sapiro, G., McMahon, K.L., de Zubicaray, G.I., Martin, N.G., Wright, M.J., Toga, A.W., Thompson, P.M., 2014. Automatic clustering and population analysis of white matter tracts using maximum density paths. Neuroimage 97, 284–295.

Qi, C.R., Su, H., Mo, K., Guibas, L.J., 2017. Pointnet: deep learning on point sets for 3d classification and segmentation. In: Proceedings of the IEEE Conference on Computer Vision and Pattern Recognition, pp. 652–660.

Reddy, C.P., Rathi, Y., 2016. Joint multi-fiber NODDI parameter estimation and tractography using the unscented information filter. Front. Neurosci. 10, 166.

Román, C., Guevara, M., Valenzuela, R., Figueroa, M., Houenou, J., Duclap, D., Poupon, C., Mangin, J.-F., Guevara, P., 2017. Clustering of whole-brain white matter short association bundles using HARDI data. Front. Neuroinform. 11, 73.

Román, C., Hernández, C., Figueroa, M., Houenou, J., Poupon, C., Mangin, J.-F., Guevara, P., 2022. Superficial white matter bundle atlas based on hierarchical fiber clustering over probabilistic tractography data. Neuroimage 262, 119550.

Román, C., López-López, N., Houenou, J., Poupon, C., Mangin, J.-F., Hernández, C., Guevara, P., 2021. Study of precentral-postcentral connections on Hcp data using probabilistic tractography and fiber clustering. In: 2021 IEEE 18th International Symposium on Biomedical Imaging (ISBI). IEEE, pp. 55–59.

Ronneberger, O., Fischer, P., Brox, T., 2015. U-Net: Convolutional Networks For Biomedical Image Segmentation, in: Medical Image Computing and Computer-Assisted Intervention – MICCAI 2015. Springer International Publishing, pp. 234–241.

Shurrab, S., Duwairi, R., 2022. Self-supervised learning methods and applications in medical imaging analysis: a survey. PeerJ Comput. Sci. 8, e1045.

Siless, V., Chang, K., Fischl, B., Yendiki, A., 2018. AnatomiCuts: hierarchical clustering of tractography streamlines based on anatomical similarity. Neuroimage 166, 32–45.

Siless, V., Davidow, J.Y., Nielsen, J., Fan, Q., Hedden, T., Hollinshead, M., Beam, E., Vidal Bustamante, C.M., Garrad, M.C., Santillana, R., Smith, E.E., Hamadeh, A., Snyder, J.,

Drews, H.H., Van Dijk, K.R.A., Sheridan, M., Somerville, L.H., Yendiki, A., 2020. Registration-free analysis of diffusion MRI tractography data across subjects through the human lifespan. Neuroimage 214, 116703.

Simonyan, K., Zisserman, A., 2014. Very deep convolutional networks for large-scale image recognition. arXiv preprint arXiv:1409.1556.

Song, H.O., Xiang, Y., Jegelka, S., Savarese, S., 2016. Deep metric learning via lifted structured feature embedding. In: 2016 IEEE Conference on Computer Vision and Pattern Recognition (CVPR), pp. 4004–4012.

Spitzer, H., Kiwitz, K., Amunts, K., Harmeling, S., Dickscheid, T., 2018. Improving Cytoarchitectonic Segmentation of Human Brain Areas with Self-supervised Siamese Networks, in: Medical Image Computing and Computer Assisted Intervention – MICCAI 2018. Springer International Publishing, pp. 663–671.

St-Onge, E., Garyfallidis, E., Collins, D.L., 2021. Fast Tractography Streamline Search, in: Computational Diffusion MRI. Springer International Publishing, pp. 82–95.

Sydnor, V.J., Rivas-Grajales, A.M., Lyall, A.E., Zhang, F., Bouix, S., Karmacharya, S., Shenton, M.E., Westin, C.-F., Makris, N., Wassermann, D., O'Donnell, L.J., Kubicki, M., 2018. A comparison of three fiber tract delineation methods and their impact on white matter analysis. Neuroimage 178, 318–331.

Tian, F., Gao, B., Cui, Q., Chen, E., Liu, T.-Y., 2014. Learning deep representations for graph clustering. In: Proceedings of the AAAI Conference on Artificial Intelligence, 28.

Tunç, B., Ingalhalikar, M., Parker, D., Lecoeur, J., Singh, N., Wolf, R.L., Macyszyn, L., Brem, S., Verma, R., 2016. Individualized map of white matter pathways: connectivity-based paradigm for neurosurgical planning. Neurosurgery 79, 568–577.

Tunç, B., Parker, W.A., Ingalhalikar, M., Verma, R., 2014. Automated tract extraction via atlas based adaptive clustering. Neuroimage 102 (Pt 2), 596–607.

Tunç, B., Smith, A.R., Wasserman, D., Pennec, X., Wells, W.M., Verma, R., Pohl, K.M., 2013. Multinomial probabilistic fiber representation for connectivity driven clustering. In: International Conference on Information Processing in Medical Imaging. Springer, pp. 730–741.

van den Oord, A., Li, Y., Vinyals, O., 2018. Representation learning with contrastive predictive coding. arXiv [cs.LG]: 1807.03748 (2018).

Van Essen, D.C., Smith, S.M., Barch, D.M., Behrens, T.E.J., Yacoub, E., Ugurbil, K., Consortium, WU-Minn HCP, 2013. The WU-Minn Human Connectome Project: an overview. Neuroimage 80, 62–79.

Vázquez, A., López-López, N., Houenou, J., Poupon, C., Mangin, J.-F., Ladra, S., Guevara, P., 2020a. Automatic group-wise whole-brain short association fiber bundle labeling based on clustering and cortical surface information. Biomed. Eng. Online 19, 42.

Vázquez, A., López-López, N., Sánchez, A., Houenou, J., Poupon, C., Mangin, J.-F., Hernández, C., Guevara, P., 2020b. FFClust: fast fiber clustering for large tractography datasets for a detailed study of brain connectivity. Neuroimage 220, 117070.

Wang, Y., Sun, Y., Liu, Z., Sarma, S.E., Bronstein, M.M., Solomon, J.M., 2019. Dynamic Graph CNN for Learning on Point Clouds. ACM Trans. Graph. 38, 1–12.

Wasserthal, J., Neher, P., Maier-Hein, K.H., 2018. TractSeg - Fast and accurate white matter tract segmentation. Neuroimage 183, 239–253.

Welling, Kipf, 2016. Semi-supervised classification with graph convolutional networks. International Conference on Learning Representations (ICLR 2017).

Wu, Y., Ahmad, S., Yap, P.-.T., 2021. Highly Reproducible Whole Brain Parcellation in Individuals via Voxel Annotation With Fiber Clusters, in: Medical Image Computing and Computer Assisted Intervention – MICCAI 2021. Springer International Publishing, pp. 477–486.

Wu, Y., Hong, Y., Ahmad, S., Lin, W., Shen, D., Yap, P.-.T., 2020. Tract Dictionary Learning For Fast and Robust Recognition of Fiber Bundles, in: Medical Image Computing and Computer Assisted Intervention – MICCAI 2020. Springer International Publishing, pp. 251–259.

Xie, J., Girshick, R., Farhadi, A., 2016. Unsupervised deep embedding for clustering analysis. In: Balcan, M.F., Weinberger, K.Q. (Eds.), Proceedings of The 33rd International Conference on Machine Learning, Proceedings of Machine Learning Research. PMLR, pp. 478–487.

Xu, C., Sun, G., Liang, R., Xu, X., 2021. Vector field streamline clustering framework for brain fiber tract segmentation. IEEE Trans. Cogn. Dev. Syst. 1.

Xu, D., Tian, Y., 2015. A comprehensive survey of clustering algorithms. Ann. Data Sci. 2, 165–193.

Xue, T., Zhang, F., Zhang, C., Chen, Y., Song, Y., Golby, A.J., Makris, N., Rathi, Y., Cai, W., O'Donnell, L.J., 2023. Superficial white matter analysis: an efficient point-cloud-based deep learning framework with supervised contrastive learning for consistent tractography parcellation across populations and dMRI acquisitions. Med. Image Anal. 85, 102759.

Xue, T., Zhang, F., Zhang, C., Chen, Y., Song, Y., Makris, N., Rathi, Y., Cai, W., O'Donnell, L.J., 2022. Supwma: consistent and efficient tractography parcellation of superficial white matter with deep learning. In: 2022 IEEE 19th International Symposium on Biomedical Imaging (ISBI). IEEE, pp. 1–5.

Xu, H., Dong, M., Lee, M.-H., OrHara, N., Asano, E., Jeong, J.-.W., 2019. Objective Detection of eloquent axonal pathways to minimize postoperative deficits in pediatric epilepsy surgery using diffusion tractography and convolutional neural networks. IEEE Trans. Med. Imaging 38, 1910–1922.

Yamada, K., Sakai, K., Akazawa, K., Yuen, S., Nishimura, T., 2009. MR tractography: a review of its clinical applications. Magn. Reson. Med. Sci. 8, 165–174.

Yeh, F.-.C., Panesar, S., Fernandes, D., Meola, A., Yoshino, M., Fernandez-Miranda, J.C., Vettel, J.M., Verstynen, T., 2018. Population-averaged atlas of the macroscale human structural connectome and its network topology. Neuroimage 178, 57–68.

Yoo, S.W., Guevara, P., Jeong, Y., Yoo, K., Shin, J.S., Mangin, J.-.F., Seong, J.-.K., 2015. An Example-Based Multi-Atlas Approach to Automatic Labeling of White Matter Tracts. PLoS One 10, e0133337.

Zekelman, L.R., Zhang, F., Makris, N., He, J., Chen, Y., Xue, T., Liera, D., Drane, D.L., Rathi, Y., Golby, A.J., O'Donnell, L.J., 2022. White matter association tracts underlying language and theory of mind: an investigation of 809 brains from the Human Connectome Project. Neuroimage 246, 118739.

Zhang, F., Cetin Karayumak, S., Hoffmann, N., Rathi, Y., Golby, A.J., O'Donnell, L.J., 2020. Deep white matter analysis (DeepWMA): fast and consistent tractography segmentation. Med. Image Anal. 65, 101761.

Zhang, F., Cetin-Karayumak, S., Pieper, S., Rathi, Y., O'Donnell, L.J., 2022a. Consistent white matter parcellation in adolescent brain cognitive development (ABCD): A ~10k Harmonized Diffusion MRI Study. Annual Meeting of the International Society for Magnetic Resonance in Medicine – ISMRM 2022.

Zhang, F., Daducci, A., He, Y., Schiavi, S., Seguin, C., Smith, R.E., Yeh, C.-.H., Zhao, T., O'Donnell, L.J., 2022b. Quantitative mapping of the brain's structural connectivity using diffusion MRI tractography: a review. Neuroimage 249, 118870.

Zhang, F., Hoffmann, N., Karayumak, S.C., Rathi, Y., Golby, A.J., O'Donnell, L.J., 2019a. Deep White Matter Analysis: fast, Consistent Tractography Segmentation Across Populations and dMRI acquisitions, in: Medical Image Computing and Computer Assisted Intervention – MICCAI 2019. Springer International Publishing, pp. 599–608.

Zhang, F., Norton, I., Cai, W., Song, Y., Wells, W.M., O'Donnell, L.J., 2017a. Comparison between two white matter segmentation strategies: an investigation into white matter segmentation consistency. In: 2017 IEEE 14th International Symposium on Biomedical Imaging (ISBI). IEEE, pp. 796–799.

Zhang, F., Savadjiev, P., Cai, W., Song, Y., Rathi, Y., Tunç, B., Parker, D., Kapur, T., Schultz, R.T., Makris, N., Verma, R., O'Donnell, L.J., 2018a. Whole brain white matter connectivity analysis using machine learning: an application to autism. Neuroimage 172, 826–837.

Zhang, F., Wu, W., Ning, L., McAnulty, G., Waber, D., Gagoski, B., Sarill, K., Hamoda, H.M., Song, Y., Cai, W., Rathi, Y., O'Donnell, L.J., 2018b. Suprathreshold fiber cluster statistics: leveraging white matter geometry to enhance tractography statistical analysis. Neuroimage 171, 341–354.

Zhang, F., Wu, Y., Norton, I., Rathi, Y., Golby, A.J., O'Donnell, L.J., 2019b. Test-retest reproducibility of white matter parcellation using diffusion MRI tractography fiber clustering. Hum. Brain Mapp. 40, 3041–3057.

Zhang, F., Wu, Y., Norton, I., Rigolo, L., Rathi, Y., Makris, N., O'Donnell, L.J., 2018c. An anatomically curated fiber clustering white matter atlas for consistent white matter tract parcellation across the lifespan. Neuroimage 179, 429–447.

Zhang, P., Wang, F., Zheng, Y., 2017b. Self supervised deep representation learning for fine-grained body part recognition. In: 2017 IEEE 14th International Symposium on Biomedical Imaging (ISBI). IEEE, pp. 578–582.

Zhang, R., Isola, P., Efros, A.A., 2016. Colorful image colorization. In: European Conference on Computer Vision (ECCV 2016). Springer, pp. 649–666.